\documentclass[letterpaper]{article} 
\usepackage{aaai24}  
\usepackage{times}  
\usepackage{helvet}  
\usepackage{courier}  
\usepackage[hyphens]{url}  
\usepackage{graphicx} 
\usepackage{natbib}  
\usepackage{caption} 
\usepackage{subcaption}
\usepackage{amsmath}
\usepackage{multirow}
\usepackage{booktabs}
\usepackage{algorithm}
\usepackage{algorithmic}

\usepackage{newfloat}
\usepackage{listings}
\DeclareCaptionStyle{ruled}{labelfont=normalfont,labelsep=colon,strut=off} 
\floatstyle{ruled}
\newfloat{listing}{tb}{lst}{}
\floatname{listing}{Listing}
\nocopyright 

\title{Deep Prompt Tuning for Graph Transformers}

\author {
    Reza Shirkavand,\textsuperscript{\rm 1}
    Heng Huang, \textsuperscript{\rm 1}
}
\affiliations {
    \textsuperscript{\rm 1} University of Maryland\\
    rezashkv@umd.edu, heng@umd.edu,
}

\begin{document}

\maketitle

\begin{abstract}
Graph transformers have gained popularity in various graph-based tasks by addressing challenges faced by traditional Graph Neural Networks. However, the quadratic complexity of self-attention operations and the extensive layering in graph transformer architectures present challenges when applying them to graph based prediction tasks. Fine-tuning, a common approach, is resource-intensive and requires storing multiple copies of large models. We propose a novel approach called deep graph prompt tuning as an alternative to fine-tuning for leveraging large graph transformer models in downstream graph based prediction tasks. Our method introduces trainable feature nodes to the graph and pre-pends task-specific tokens to the graph transformer, enhancing the model's expressive power. By freezing the pre-trained parameters and only updating the added tokens, our approach reduces the number of free parameters and eliminates the need for multiple model copies, making it suitable for small datasets and scalable to large graphs. Through extensive experiments on various-sized datasets, we demonstrate that deep graph prompt tuning achieves comparable or even superior performance to fine-tuning, despite utilizing significantly fewer task-specific parameters. Our contributions include the introduction of prompt tuning for graph transformers, its application to both graph transformers and message passing graph neural networks, improved efficiency and resource utilization, and compelling experimental results. 
This work brings attention to a promising approach to leverage pre-trained models in graph based prediction tasks and offers new opportunities for exploring and advancing graph representation learning.
\end{abstract}

\section{Introduction}
\par
The remarkable success of transformers in Natural Language Processing \cite{vaswani2017attention} and Computer Vision \cite{dosovitskiy2021vit} has led to their increasing popularity in graph applications. In recent years, graph transformers have been widely adopted in various graph-based tasks \cite{dwivedi2020gt, kruezer2021san, ying2021graphormer, chen2022sat, ladislav2022graphgps}. Graph transformers address key challenges faced by traditional Graph Neural Networks (GNNs), such as limited expressiveness \cite{wang22spectralexpress,feng22mpgexpress}, over-smoothing \cite{chen20oversmoothing} and over-squashing \cite{alon21oversquashing}, by leveraging highly expressive global self-attention modules. Instead of introducing graph structure bias into intermediate layers, graph transformers use encoded structural and positional information of the graph into the input node features \cite{chen2022sat, kruezer2021san, ying2021graphormer}.

\par
The quadratic complexity of the self-attention operation, combined with the extensive layering in graph transformer architectures, presents a significant challenge when applying them to graph prediction tasks. This challenge arises from the potential overfitting to small datasets, and even reduced-parameter versions may lack the necessary representational richness for complex graph datasets, such as those involved in molecular property prediction. To address this issue, a viable solution is to adopt a similar approach as in the application of large language models to downstream NLP tasks \cite{radford2019language,bert}. This involves pre-training large graph transformer models on extensive datasets and subsequently fine-tuning them on smaller datasets.

\par
While fine-tuning can yield satisfactory results \cite{fine-tuning}, it presents a resource-intensive approach due to the need to update the entire parameter set of the large graph transformer model. Furthermore, fine-tuning necessitates storing multiple copies of the same large model for different downstream tasks, which can be especially challenging and even prohibitive on smaller devices. In contrast to fine-tuning, prompt tuning, originally proposed for natural language processing (NLP) tasks, offers an alternative solution. Prompt tuning involves freezing all parameters of a pre-trained model and only updating either discrete \cite{prompt1,prompt2} or continuous \cite{lia21prefixtuning,lester21promptnlp1,promptnlp2,p-tuning-v2} lightweight tokens that are added to the inputs. Despite the success of prompt tuning in the field of NLP, the application of this technique to graph transformers has yet to be experimentally explored.

\par
Motivated by the concept of prompt tuning, we introduce a novel approach called \emph{Deep Graph Prompt Tuning (DeepGPT)}, which serves as an alternative to fine-tuning for leveraging large graph transformer models in downstream graph prediction tasks. Fig. \ref{fig:method} illustrates the setup of a typical downstream graph prediction task, where an input graph is transformed into a final graph representation vector as the output. Assuming we have access to a pre-trained large graph transformer network trained on a different dataset, DeepGPT begins by adding a continuous task-specific graph \emph{prompt token} to the feature vectors of all nodes within the input graph. Subsequently, the modified input graph is fed into the graph transformer. Although this approach focuses on modifying the graph input level only, it theoretically allows for approximating various combinations of node-level, edge-level, and graph-level transformations within the architectures of existing pre-trained GNN models (\cite{fang22graphprompt}. Additionally, we pre-pend continuous layer-specific task-specific \emph{prefix tokens} to the embeddings of each layer in the graph transformer network. This enables the self-attention module of each transformer layer to attend to the trainable task-specific tokens as if they were originally part of the sequence of node embeddings. It also increases the expressive power of our DeepGPT method. While our approach is specifically designed for graph transformers with global attention modules, it can be viewed as introducing a set of new nodes with trainable features to the graph and connecting them to all pre-existing nodes in the context of Message Passing Graph Neural Networks (MPGNNs).

\par
In contrast to fine-tuning, our approach involves freezing all parameters of the pre-trained graph transformer and exclusively tuning the task-specific tokens that are added. This allows for the practicality of storing a single copy of the pre-trained transformer architecture alongside the task-specific continuous prompts for different downstream tasks. Through extensive experiments conducted on various-sized OGB and Moleculenet datasets, we have demonstrated that our approach delivers comparable, and in some cases even superior, performance compared to fine-tuning, despite utilizing fewer than 0.5\% of task-specific parameters. This significant reduction in the number of free parameters proves advantageous for small datasets as it mitigates the risk of overfitting. Furthermore, our method enables the pre-trained graph transformer to scale effectively to large graphs, as the required computational resources are substantially reduced.

\par
Our main contributions can be summarized as follows:
\begin{enumerate}
    \item 
    We propose a novel prompt tuning method specifically designed for graph transformers, which, to the best of our knowledge, is the first of its kind in the field.
    \item  Our approach, designed for graph transformers with global attention modules, is equivalent to introducing additional trainable feature nodes connected to existing nodes within the context of MPGNNs, making it applicable to both Graph Transformers and MPGNNs.
    \item 
    Our approach significantly improves the efficiency and resource utilization of tuning a pre-trained graph transformer for downstream tasks. It eliminates the need to store separate copies of a large model for different graph prediction tasks, making it a more streamlined solution. Additionally, our method is well-suited for both small datasets and scales effectively to handle datasets with large graphs, addressing the limitations of graph transformers in dealing with such scenarios.
    \item 
    Our extensive experiments demonstrate that our method achieves performance that is on par with, and at times surpasses, fine-tuning, despite utilizing significantly fewer task-specific parameters. Moreover, our method demonstrates superiority over lightweight fine-tuning, where only the classification head is updated. 

\end{enumerate}

\section{Related Work}
\subsection{Fine-tuning Strategies for Graph Neural Networks}
\label{subsec: fine-tune gnn}
Several approaches have been explored in fine-tuning graph neural networks. For instance, ChemBERTa \cite{chemberta} adopts an NLP self-supervised pre-training strategy inspired by RoBERTa \cite{roberta}, where a portion of tokens in the SMILES string representation of PUBCHEM graphs are masked, and the model predicts them from other tokens. \cite{MGSSL} propose a motif-based generative pre-training framework, training GNNs to make topological and label predictions. Another work by \cite{selfsupervised-gt} introduces a molecular graph pre-training model with self-supervised tasks at the node, edge, and graph levels, including contextual property prediction and graph-level motif prediction. \cite{han2022kpgt} design a generative self-supervised strategy that leverages molecular descriptors and fingerprints to guide the model in capturing structural and semantic information from large-scale unlabeled molecular graph datasets. Graph Contrastive Learning (GCL) objectives have also been utilized in several studies \cite{GraphCL, GCL2, DGI} to enhance the pre-training process. Additionally, \cite{ying2021graphormer} pre-train their proposed Graphormer model on large-scale OGB datasets to capture rich representations, followed by fine-tuning for specific target tasks.

\subsection{Prompt Tuning for Graph Neural Networks}
\label{subsec: prompt-tuning gnn}
Although there has been an extensive amount of research into both discrete \cite{sentiment-analysis,text-classification,autoprompt} and continuous \cite{p-tuning-v2, promptnlp2, lester21promptnlp1} prompt tuning for NLP, the application of prompts to the graph domain is relatively unexplored. \cite{sun22gppt} propose a graph prompting function to transform a downstream node prediction task to an edge prediction task similar to the pretext masked-edge prediction used to pre-train a large GNN. \cite{fang22graphprompt} propose Graph Prompt (GP) as a universal graph prompt tuning method to existing pre-trained GNNs. They also introduce Graph Prompt Features (GPF) as a concrete instance of GP. GPF is a trainable token added to the node features of the input graph and updated during the tuning of the pre-trained model on the downstream task.

\section{Preliminary}
In this section, we will cover the fundamentals of graph neural networks (GNNs) and graph transformers, as well as provide an overview of the ``pre-train, fine-tune" paradigm.

\subsection{Graph Neural Networks}

Let us consider an undirected graph $G=(V, E)$, where $V = \{1, \cdots, n\}$ represents the set of nodes and $E$ represents the set of edges. We assume that each node $i$ is associated with a d-dimensional feature vector $x_{i} \in R^{d}$ for $i = 1, \cdots, n$. Graph Neural Networks (GNNs) employ a message-passing strategy \cite{gilmer2017messagepassing} to learn node representations by iteratively aggregating the representations of neighboring nodes. Formally, the representation of node $i$ at the $k$-th layer is denoted as $h_{i}^{k}$, with $h_{i}^{0}=x_{i}$. The aggregation and combination operation is defined as follows:

\begin{equation}
h_{i}^{k} = \text{AGG-COMB} ({h_{j}^{k-1}: j \in N(i) \cup i};\theta^{k})
\end{equation}

Here, $N(i)$ represents the set of adjacent nodes to node $i$, and $\theta^{k}$ is the parameter set of the $k$-th layer. The AGG-COMB operation collects the embeddings of neighboring nodes and combines them using sum, mean, or max functions to generate the embeddings for node $i$. In graph representation learning tasks, a final READOUT function is employed to combine the node representations of a $K$-layer GNN into the graph embedding $h_{G}$:

\begin{equation}
h_{G} = \text{READOUT} ({h_{i}^{K}: i = 1, \cdots, n})
\end{equation}

Graph Neural Networks have demonstrated superior performance in various graph prediction tasks, including node-level \cite{gao2018nodelevel}, edge-level \cite{li21edgelevel}, and graph-level \cite{hao20graphlevel} tasks.

\subsection{Graph Transformers}

Unlike conventional GNNs, Transformers \cite{vaswani2017attention} do not explicitly utilize the graph structure to learn representations. Instead, they treat the graph as a set of nodes and infer similarities between nodes by employing self-attention mechanisms on node features. A typical transformer block consists of a multi-head attention module followed by a feed-forward network. The input node feature matrix $X$ is linearly projected into Query (Q), Key (K), and Value (V) matrices, represented as $Q = XW_{Q}$, $K = XW_{K}$, and $V = XW_{V}$, respectively. The self-attention activations are computed as follows:

\begin{equation}
Attn(X) = \text{softmax}\left(\frac{QK^{T}}{\sqrt{d_Q}}\right)V
\end{equation}

Here, $d_Q$ represents the dimension of $Q$. The output of the transformer layer is obtained by applying a feed-forward network (FNN) to the sum of the input node features and the self-attention activations:

\begin{equation}
X = \text{FFN}(X + Attn(X))
\end{equation}

Multiple transformer layers are usually stacked together to form a transformer network. Since self-attention is a permutation-invariant operator, the transformer produces the same result regardless of changes in the graph structure, as long as the node features remain unchanged. To incorporate the structural information of the graph into the transformer architecture, effective absolute encodings are designed. These encodings are structural and positional representations of the graph that are added or concatenated to the input features to enhance the expressiveness and generalization capability of Graph Transformers. Examples of absolute encodings include Centrality Encoding \cite{ying2021graphormer}, RWPE \cite{dwivedi2022rwpe}, and Laplacian Positional Encoding \cite{kruezer2021san}.

\subsection{Pre-training and Fine-tuning of GNNs}

Supervised learning of node and graph representations often requires a large amount of annotated data, which can be challenging to obtain. Consequently, the ``pre-train, fine-tune" paradigm has gained significant attention. In this paradigm, neural networks are pre-trained on pretext tasks, such as Contrastive Learning \cite{hu2020generativepretraining, qui2020contrastive-1, subramonian2021contrastive2}, and then fine-tuned on downstream tasks. The process involves training a graph neural network $f_{\theta}$ on a pre-training dataset $D_{pt}$ by minimizing the pre-training loss $L_{pt}$, resulting in parameter set $\theta_{pt}$. The downstream model parameters are initialized using $\theta_{init} = \theta_{pt}$, and the prediction head of the GNN is replaced. Fine-tuning is performed by optimizing the downstream loss $L$ on the downstream dataset $D$ using the whole parameter set of $f$ or a subset of it:

\begin{equation}
\min_{\theta, \psi} L(p_{\psi}(f_{\theta}(x_{i})), y_{i})
\end{equation}

Here, $p_{\psi}$ denotes the new prediction head of the network, and $y_i$ corresponds to the ground truth of $x_i$.

\section{Methodology}
\begin{figure*}[t]
  \centering
  \includegraphics[width=0.8\textwidth]{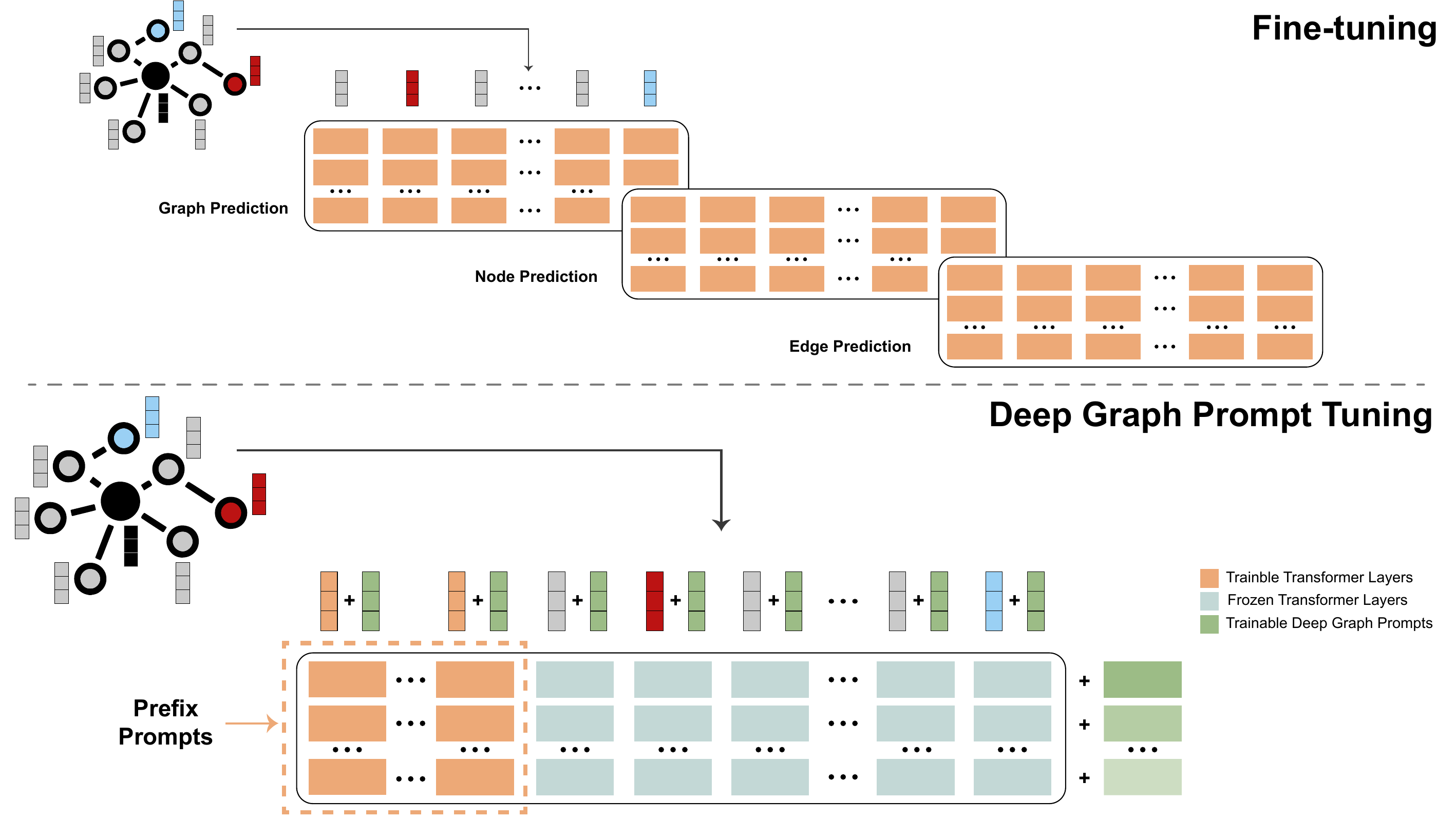}
  \caption{Overview of our proposed deep graph prompt tuning framework. \textbf{Top}: Fine-tuning requires storing separate copies of the pre-trained graph transformer for each downstream task and updating all parameters of the model. This approach is memory-intensive, time-consuming, and cost-inefficient. \textbf{Bottom}: our proposed graph prompt tuning method. We introduce graph prompt tokens to the input graph representations and each transformer layer activation. Additionally, we pre-pend prefix tokens to all transformer layers. By freezing most of the pre-trained model's parameters and only updating the concatenated and added prompt tokens, we can fine-tune a large transformer on different downstream datasets while only storing the prompt tokens, minimizing memory requirements.}
  \label{fig:method}
\end{figure*}

The ``unsupervised pre-train, fine-tune" framework faces a significant challenge due to the training objective gap between pretext and downstream tasks. This issue becomes more pronounced when dealing with large neural networks like transformers, as it requires storing separate copies of the pre-trained network for each downstream task, incurring high costs. Moreover, fine-tuning these massive models on large downstream datasets is time-consuming. To address these limitations, we draw inspiration from the success of continuous prompt-based learning in NLP \cite{promptnlp2,lester21promptnlp1} and propose deep prompt tuning for Graph Transformers as an alternative approach for node/graph classification tasks. The framework, depicted in Figure \ref{fig:method}, leverages continuous prompts to guide the graph transformer's learning process, reducing the need for separate network copies and significantly improving efficiency.

Based on intuition from prompt tuning, we believe that providing a proper context in the form of graph embeddings can guide a graph transformer without altering its parameters. By incorporating relevant graph embeddings as context, the graph transformer can achieve higher accuracy in graph prediction tasks. This approach extends the concept of prompting beyond adding single nodes or edges and aims to find a context that influences the transformer's encoding of graphs and the generation of predictions. Rather than discrete alterations to the graph \cite{sun22gppt}, we propose optimizing continuous graph embeddings as context, which can effectively propagate throughout the graph transformer's  layers, striking a balance between expressiveness and computational feasibility. 

Our approach entails transforming both the input graph and the pre-trained graph transformer model. We combine graph prompts with deep prefix prompts. This involves incorporating graph prompt tokens at the input of the graph transformer and pre-pending prefix tokens to each transformer layer and to direct the graph transformer in solving downstream node/graph classification tasks by fine-tuning these task-specific tokens. We demonstrate the superiority of this approach over using solely graph prompts or prefix tokens, despite utilizing the same number of parameters. Through our experiments, we showcase the acceptable, and at times even superior, performance of our approach with significantly fewer parameters compared to traditional fine-tuning, where all network parameters are adjusted, or lightweight fine-tuning, which involves freezing the pre-trained model's backbone and training only the prediction head.

\subsection{Graph Prompt Tokens}
In our prompt tuning method, the first step involves incorporating graph prompt tokens into the input graph nodes. For a $d$-dimensional node feature vector $x_{i} \in R^{d}$, we introduce a trainable $d$-dimensional prompt token $p \in R^{d}$ specific to the task, which is added to each node of the input graph. This results in a modified feature matrix $X^{(p)}$ derived from the original feature matrix $X_i$ as follows:
\begin{equation}
x_{i}^{(p)} = x_{i} + p
\end{equation}

Prior research by \cite{fang22graphprompt} has demonstrated that such prompt tokens possess the theoretical capability to approximate a wide range of complex node-level, edge-level, and graph-level transformations on most existing Graph Neural Networks (GNNs) under two conditions: utilizing a Graph Convolutional Network (GCN) as the underlying architecture and employing sum/mean as the READOUT operator. Formally, for an input graph $G(X, A)$, with $X$ representing the node features and $A$ denoting the adjacency matrix, given an arbitrary transformation $g$, and a pre-trained GNN $f$, there exists a corresponding prompt token $p^*$ satisfying:
\begin{equation}
f(X^{p^{*}}, A) = f(g(X, A))
\end{equation}

Although the theoretical foundation of the graph prompt approach primarily focuses on Convolutional Graph Neural Networks (GNNs), we empirically demonstrate that the benefits of graph prompts extend to graph transformers as well. The shared principles between these two models, such as alignment with the READOUT operator and the introduction of task-specific information, contribute to the improved performance of graph transformers when employing the graph prompt technique.

\subsection{Graph Transformer Prefix Tokens}
The second component of our approach involves the inclusion of prefix tokens at the beginning of the embeddings after each layer of the frozen transformer architecture, resulting in updated embeddings. Let $PS_{idx}$ denote the sequence of prefix tokens, where $p=|PS_{idx}|$ represents the length of the sequence. Assuming the dimension of the transformer embeddings is $d$, and denoting the original transformer layer embeddings as $E$, we utilize a soft prompt matrix $P \in R^{p \times d}$ to generate the new embeddings $E^{*}$ according to the following rule:
\begin{equation}
    E^{*}[i,:] = 
    \begin{cases}
    P[i, :] & i \in PS_{idx}\\
    E[i, :] & otherwise
    \end{cases}
\end{equation}

The resulting embedding matrix is then passed as input to the subsequent transformer layer. Importantly, while the graph prompt tokens vary with each layer, the tokens in earlier layers still have an impact on the embeddings of later layers. This is due to the influence of the prompt token on the input embeddings through the self-attention operation, enabling these tokens to propagate throughout the entire transformer architecture.

\subsection{Deep Graph Transformer Prompt Tuning}
The shallow approach of simply appending or adding prompt tokens to the input graph \cite{lia21prefixtuning} encounters two significant challenges. Firstly, the limited number of trainable parameters leads to unstable training and unsatisfactory results. Secondly, adding prompts only to the input graph has minimal impact on the deeper layers of the transformer. To address these limitations, we propose the Deep Graph Prompt Tuning (DeepGPT) technique, which involves adding a prompt token to the input the graph transformer and pre-pending prefix prompts to each transformer layer embedding.

Assuming a pre-trained graph transformer $f_\theta$, DeepGPT transformation $T_{\phi}$, and a downstream dataset $D={(G_{i}, y_{i})}{i=1}^{n}$, we fine-tune the trainable parameter set $\phi$ while keeping the parameters $\theta$ of the pre-trained graph transformer frozen. This fine-tuning process aims to minimize the downstream loss $L$, such as the binary cross-entropy loss, given by the following optimization objective:
\begin{equation}
\min{\phi} \sum_{i=1}^{n} L(f_\theta(T_{\phi}(G_{i})), y_{i})
\end{equation}

\subsection{Relation to Conventional Graph Neural Networks}
Although our work primarily focuses on graph transformer architectures, the construct of prompt tokens we propose is applicable universally. The graph prompt tokens can be incorporated into various graph neural network models without assuming a specific structure for the model $f$. Similarly, in the context of Message Passing Graph Neural Networks (MPGNNs), the prefix tokens of the transformer architecture can be replaced by adding a set of new trainable nodes to the graph and connecting them with edges to all existing nodes, thereby achieving similar effects.

\section{Experiments }
\subsection{Experimental Settings}
\subsubsection{Datasets}
The graph transformer models in this study were pre-trained on the OGB-LSC \cite{hu2021ogblsc} quantum chemistry regression dataset, known as PCQM4Mv2. After pre-training, we assess the efficacy of our approach using two established sets of machine learning datasets focused on molecular graphs: the Open Graph Benchmark (OGB) \cite{hu2020ogb}, which encompasses datasets of diverse sizes, covering a wide range of realistic tasks, and the Moleculenet Benchmark \cite{wu2018moleculenet}, which comprises datasets for predicting molecular properties. These datasets include regression, single-label binary classification, and multi-label classification tasks, spanning various domains (For further details, please refer to the Appendix).

\subsubsection{Graph Transformer Architectures}
In our study, we investigate three graph transformer architectures that address the key challenge of incorporating structure-awareness into the attention mechanism when graphs are used as input. Firstly, \cite{ying2021graphormer} present Graphormer, which modifies the attention mechanism itself to integrate structural information. Secondly, \cite{ladislav2022graphgps} propose General Powerful and Scalable Graph Transformer (GraphGPS), which employs hybrid architectures, integrating GNNs. Thirdly, \cite{han2022kpgt} introduce Line Graph Transformer (LiGhT), which utilizes positional encodings for graphs. While all three architectures share the self-attention module at their core, they exhibit fundamental differences in overall design, structure, and the usage of positional and structural encodings.

We conducted experiments with two different settings of the GraphGPS network: one using regular self-attention transformers (GraphGPS small), and the other utilizing Performer \cite{performer} modules (GraphGPS large), which are Transformer architectures capable of accurately estimating regular (softmax) full-rank-attention Transformers. The performer modules achieve this accuracy while employing linear space and time complexity instead of the quadratic complexity found in traditional methods. This allowed us to demonstrate the effectiveness of our proposed method on different self-attention modules. The choice of network is explicitly indicated in the result tables.

\subsubsection{Message-Passing GNN Baselines}
To further evaluate the effectiveness of our DeepGPT framework, we present the performance results of popular Message-Passing Graph Neural Networks (MPGNN), namely GatedGCN \cite{GatedGCN}, GINE \cite{GINE}, and PNA \cite{PNA}.

\subsubsection{Training Details}
We employ the AdamW \cite{adamw} optimizer across all architectures and datasets for both pre-training and evaluation. We utilize a learning rate scheduling with a warm-up stage followed by a cosine decay regiment. We tune a hyper-parameter set including learning rate, weight decay, and the Prefix Token size $|PS|$. We perform a 5-fold cross validation for all graph transformer experiments to have a fair evaluation. All experiments are are conducted on a Lambda machine with 8 NVIDIA RTX A6000 GPUs. (For all details, see Appendix).

\subsection{Results}
Tables \ref{tab:moleculenet-classification} and \ref{tab:moleculenet-regression} demonstrate the performance of our proposed method on the classification and regression benchmarks, respectively. The performance of DeepGPT on the OGB benchmarks is presented in Table \ref{tab:ogb-classification}. We compare DeepGPT to Fine-tuning as well as MPGNN baselines on molecular graph tasks.

\begin{table*}[ht]
\centering
\caption{Comparison of the performance of DeepGPT, Fine Tuning (FT) and MPGNN baselins on classification benchmarks. Entries marked with an asterisk ($\ast$) indicate the use of Performer self-attention. Results represent the mean and standard deviation from 5-fold cross-validation.}
\label{tab:moleculenet-classification}
\resizebox{\textwidth}{!}{%
\begin{tabular}{@{}lccccccccc@{}}
\toprule
\multirow{2}{*}{Model} & \multirow{2}{*}{\#Param.} & \multicolumn{8}{c}{Classification Dataset ($\uparrow$)}               \\ \cmidrule(l){3-10} 
                                        &     & BACE & BBBP$^\ast$ & ClinTox$^\ast$ & Estrogen$^\ast$ & MetStab$^\ast$ & SIDER & ToxCast$^\ast$ & Tox21 \\ \midrule
Graphormer \cite{ying2021graphormer} FT & 48M  & 0.704 ± 0.018 & 0.778 ± 0.043 & 0.839 ± 0.114 & 0.673 ± 0.034 & 0.607 ± 0.015 & 0.550 ± 0.007 & 0.665 ± 0.032 & 0.582 ± 0.051 \\
Graphormer DeepGPT                      & 100K & 0.883 ± 0.022 & 0.914 ± 0.016 & 0.884 ± 0.030 & 0.941 ± 0.006 & 0.871 ± 0.004 & 0.646 ± 0.003 & 0.725 ± 0.011 & 0.813 ± 0.010\\ \midrule
GraphGPS \cite{ladislav2022graphgps} FT &  14M ($\ast$: 103M)  & 0.898 ± 0.020 & 0.888 ± 0.015 & 0.910 ± 0.046 & 0.933 ± 0.014 & 0.897 ± 0.025 & 0.610 ± 0.005 & 0.722 ± 0.011 & 0.832 ± 0.009\\
GraphGPS DeepGPT                        &  50K ($\ast$: 500K)  & 0.892 ± 0.022 & 0.901 ± 0.015 & 0.907 ± 0.044 & 0.944 ± 0.010 & 0.899 ± 0.019 & 0.609 ± 0.020 & 0.735 ± 0.005 & 0.832 ± 0.012 \\ \midrule
LiGhT \cite{han2022kpgt} FT             &  90M & 0.880 ± 0.012 & 0.902 ± 0.021 & 0.857 ± 0.035 & 0.942 ± 0.012 & 0.902 ± 0.012 & 0.670 ± 0.005 & 0.745 ± 0.005 & 0.844 ± 0.004\\
LiGhT DeepGPT                           & 370K & 0.873 ± 0.020 & 0.917 ± 0.012 & 0.862 ± 0.056 & 0.950 ± 0.010 & 0.912 ± 0.011 & 0.671 ± 0.011 & 0.757 ± 0.011 & 0.843 ± 0.004\\ \midrule \midrule
GatedGCN \cite{GatedGCN}                & 2.8M & 0.833 ± 0.013 & 0.887 ± 0.025 & 0.893 ± 0.041 & 0.919 ± 0.008 & 0.848 ± 0.018 & 0.599 ± 0.015 & 0.683 ± 0.005 & 0.807 ± 0.011\\
GINE \cite{GINE}                        & 1.2M & 0.599 ± 0.045 & 0.613 ± 0.024 & 0.559 ± 0.044 & 0.492 ± 0.024 & 0.540 ± 0.016 & 0.584 ± 0.025 & 0.629 ± 0.025 & 0.714 ± 0.024\\
PNA \cite{PNA}                          & 1.8M & 0.845 ± 0.021 & 0.903 ± 0.018 & 0.867 ± 0.011 & 0.927 ± 0.009 & 0.738 ± 0.024 & 0.583 ± 0.012 & 0.673 ± 0.008 & 0.793 ± 0.015\\ \bottomrule
\end{tabular}%
}
\end{table*}

\begin{table}[ht]
\centering
\caption{Comparison of the performance of DeepGPT, Fine Tuning (FT) and MPGNN baselins on Moleculenet regression benchmarks.}
\label{tab:moleculenet-regression}
\resizebox{\columnwidth}{!}{%
\begin{tabular}{@{}lccc@{}}
\toprule
\multirow{2}{*}{Model}  & \multicolumn{3}{c}{Moleculenet Regression Dataset ($\downarrow$)} \\ \cmidrule(l){2-4} 
                      & FreeSolv      & ESOL          & Lipo \\ \midrule
Graphormer  FT        & 1.680 ± 0.013 & 0.925 ± 0.061 & 0.909 ± 0.237 \\
Graphormer DeepGPT    & 1.668 ± 0.114 & 0.943 ± 0.024 & 0.834 ± 0.046 \\ \midrule
GraphGPS FT           & 1.380 ± 0.234 & 0.772 ± 0.102 & 0.673 ± 0.022 \\
GraphGPS DeepGPT      & 1.415 ± 0.254 & 0.685 ± 0.130 & 0.528 ± 0.056 \\ \midrule
LiGhT FT              & 0.955 ± 0.051 & 0.565 ± 0.041 & 0.530 ± 0.019 \\
LiGhT DeepGPT         & 0.983 ± 0.079 & 0.579 ± 0.038 & 0.548 ± 0.028 \\ \midrule \midrule
GatedGCN              & 2.685 ± 1.766 & 1.373 ± 0.219 & 0.568 ± 0.065 \\
GINE                  & 6.971 ± 3.892 & 3.075 ± 1.034 & 1.440 ± 0.027 \\
PNA                   & 2.479 ± 0.556 & 1.463 ± 0.341 & 0.684 ± 0.024 \\ \bottomrule

\end{tabular}%
}
\end{table}

\begin{table}[ht]
\centering
\caption{Comparison of the performance of DeepGPT, Fine Tuning (FT) and MPGNN baselins on OGB classification benchmark.}
\label{tab:ogb-classification}
\resizebox{0.8\columnwidth}{!}{%
\begin{tabular}{@{}lcc@{}}
\toprule
\multirow{2}{*}{Model} & \multicolumn{2}{c}{OGB Classification Dataset ($\uparrow$)} \\ \cmidrule(l){2-3} 
                     & MOLHIV (AUROC) & MOLPCBA (AP)   \\ \midrule
Graphormer FT        & 0.805 ± 0.005  & 0.313 ± 0.003 \\
Graphormer DeepGPT   & 0.804 ± 0.021  & 0.289 ± 0.009 \\ \midrule
GraphGPS FT          & 0.806 ± 0.007  & 0.301 ± 0.013  \\
GraphGPS DeepGPT     & 0.801 ± 0.015  & 0.297 ± 0.020  \\ \midrule
LiGhT FT             & 0.787 ± 0.008  & 0.295 ± 0.006  \\
LiGhT DeepGPT        & 0.799 ± 0.010  & 0.270 ± 0.007           \\ \midrule \midrule
GatedGCN             & 0.809 ± 0.016  & 0.264 ± 0.021  \\ 
GINE                 & 0.679 ± 0.055  &   -            \\ 
PNA                  & 0.782 ± 0.013  & 0.257 ± 0.006 \\ \bottomrule
\end{tabular}%
}
\end{table}

\subsubsection{DeepGPT Across Tasks and  Model Scales}
Based on the results, we observe that DeepGPT exhibits comparable performance and, in certain cases, even outperforms fine-tuning across various tasks. Table \ref{tab:gt-sizes} (Appendix) displays the sizes of the graph transformer architectures utilized in our experiments. These results demonstrate that our method is adaptable and effective for a diverse range of graph transformer architectures, irrespective of their sizes.

\subsubsection{DeepGPT Across Dataset Scales}
It is important to highlight that the downstream datasets used in this study consist of varying dataset sizes and graph sizes, as depicted in Figure \ref{fig:datasets} (Appendix). The results demonstrate the versatility of DeepGPT, as it proves to be effective for small datasets, while also scaling well to larger datasets. Another noteworthy aspect of graph transformer prompt tuning is its ability to address the limitations of many graph transformer models, which are often unsuitable for processing large graphs. By reducing the overhead associated with fine-tuning such models on large graphs, we achieve satisfactory results compared to fine-tuning.

\subsubsection{Benchmarking DeepGPT}
The results obtained from our experiments provide compelling evidence of DeepGPF's remarkable performance, surpassing all MPGNN baselines across nearly every task evaluated. Furthermore, the findings offer a glimpse into the immense potential of prompt tuning large graph transformer models, suggesting that they have the capacity gain wider adoption in graph prediction tasks.

\subsubsection{Convergence Speed}
\begin{figure}[ht]
  \begin{subfigure}{0.45\columnwidth}
        \centering
        \includegraphics[width=\textwidth]{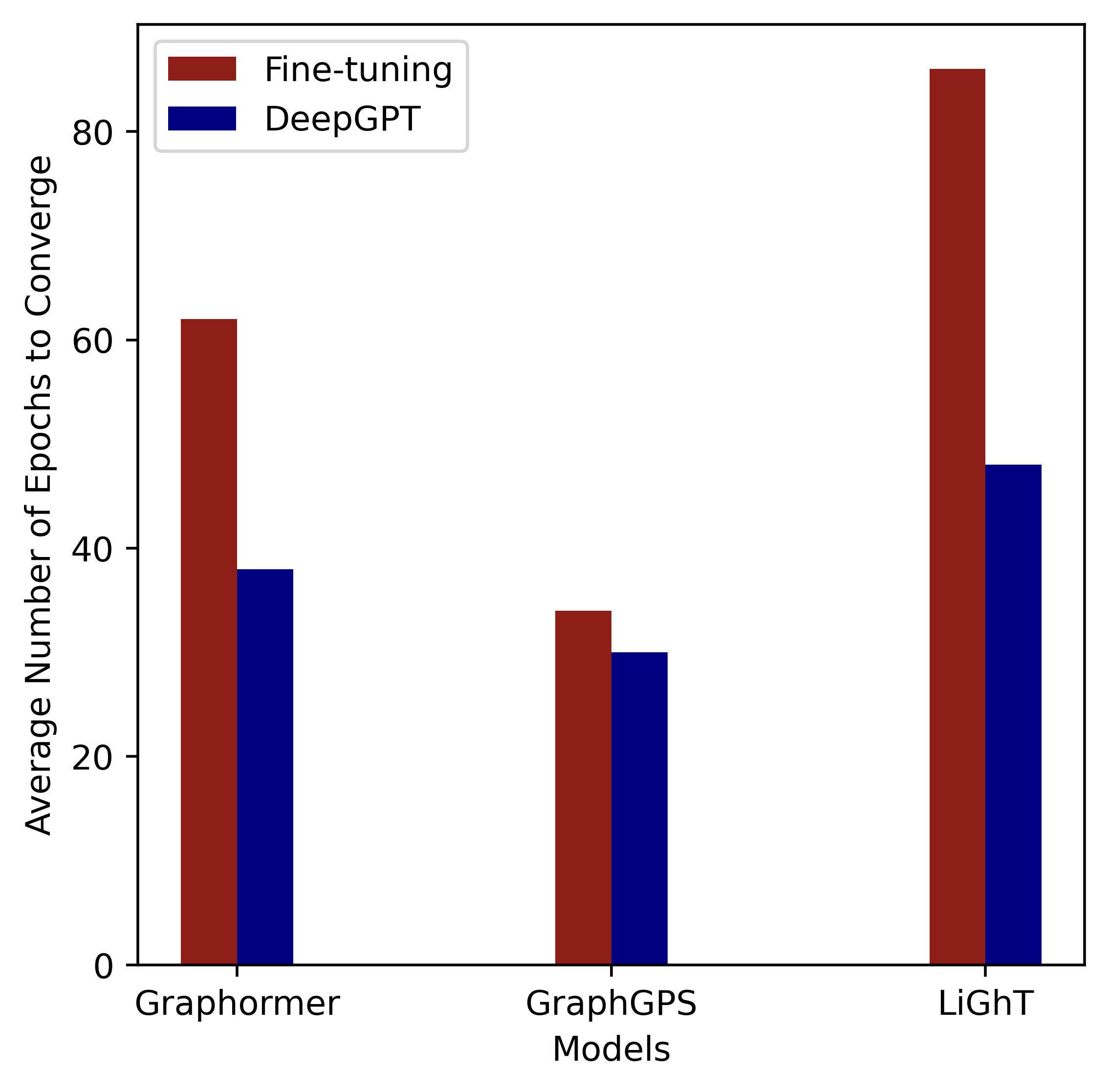}
        \caption{Average number of epochs}
        \label{fig:convergence:a}    
  \end{subfigure}
  \hfill
  \begin{subfigure}{0.45\columnwidth}
      \centering
      \includegraphics[width=\textwidth]{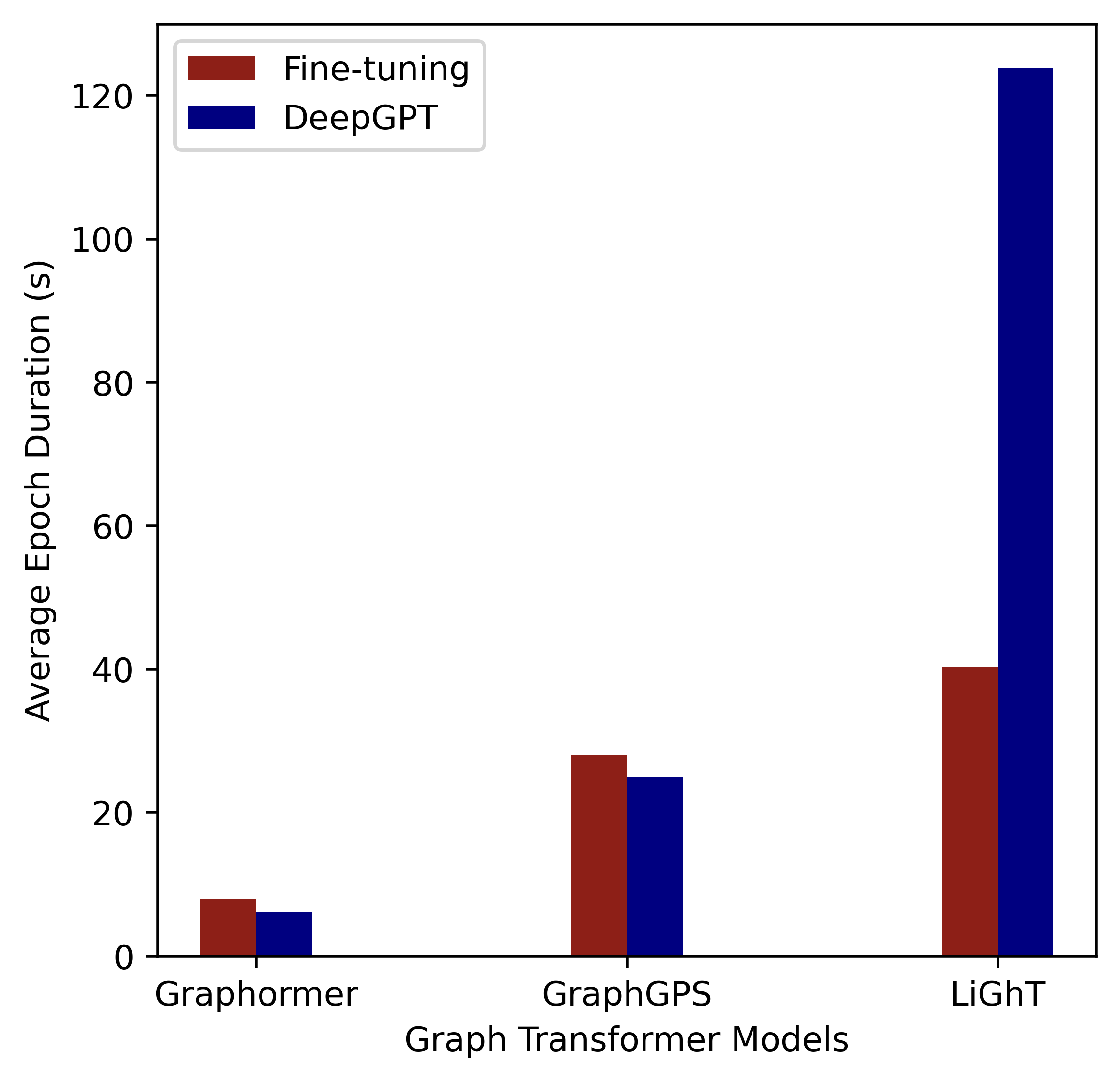}
      \caption{Average epoch duration}
      \label{fig:convergence:b}  
  \end{subfigure}
  \caption{Comparison of convergence speed of DeepGPT and FT. DeepGPT converges faster and decreases training and inference epoch duration.
  }
\label{fig:convergence}  
\end{figure}

Overall, DeepGPT exhibits faster convergence during model training. Additionally, each DeepGPT epoch is usually quicker than fine-tuning (Figure \ref{fig:convergence}). The increased average epoch duration of DeepGPT on LiGhT is attributed to the implementation of the original paper, which requires a loop for adding prompt tokens, impacting the overall speed. 

\subsection{Ablation Studies}
\subsubsection{Prompt Tokens Contributions}
Table \ref{tab:ablation-tune} demonstrates the impact of adding graph prompt tokens to deep prefix tuning. Generally, this addition leads to performance improvements even with the same number of parameters. The addition of graph prompt tokens alone has not been  experimented with due to the limited number of parameters, which leads to an insignificant impact during model training. 
Additionally, DeepGPT outperforms lightweight tuning, where only the classification head of the pre-trained model is modified.

\begin{table}[ht]
\centering
\caption{Impact of adding graph prompt tokens to deep prefix tuning.}
\label{tab:ablation-tune}
\resizebox{\columnwidth}{!}{%
\begin{tabular}{@{}lccccccccc@{}}
\toprule
\multirow{2}{*}{Model} & \multirow{2}{*}{\#Param.} & \multicolumn{4}{c}{Dataset ($\uparrow$)} \\ \cmidrule(l){3-6} 
                              &                    &      BACE     &     Clintox $^\ast$   &    Metstab $^\ast$   &     SIDER     \\ \midrule
Graphormer Lightweight Tuning & 600K               & 0.847 ± 0.013 & 0.766 ± 0.082 & 0.863 ± 0.021 & 0.625 ± 0.017 \\ 
Graphormer Deep Prefix Tuning & 100K               & 0.863 ± 0.013 & 0.860 ± 0.062 & 0.883 ± 0.014 & 0.584 ± 0.009 \\ 
Graphormer DeepGPT            & 100K               & 0.883 ± 0.022 & 0.884 ± 0.030 & 0.871 ± 0.004 & 0.646 ± 0.003 \\ \midrule
GraphGPS Lightweight Tuning   & 500K               & 0.819 ± 0.010 & 0.692 ± 0.107 & 0.673 ± 0.031 & 0.541 ± 0.015 \\
GraphGPS Deep Prefix Tuning   & 50K ($^\ast$:500K) & 0.854 ± 0.016 & 0.866 ± 0.059 & 0.841 ± 0.014 & 0.598 ± 0.019 \\
GraphGPS DeepGPT              & 50K ($^\ast$:500K) & 0.892 ± 0.022 & 0.907 ± 0.044 & 0.899 ± 0.019 & 0.609 ± 0.020 \\ \midrule
\end{tabular}%
}
\end{table}

\subsubsection{Prompt Depth}
\begin{figure}[ht]
  \begin{subfigure}{0.48\columnwidth}
        \centering
        \includegraphics[width=\textwidth]{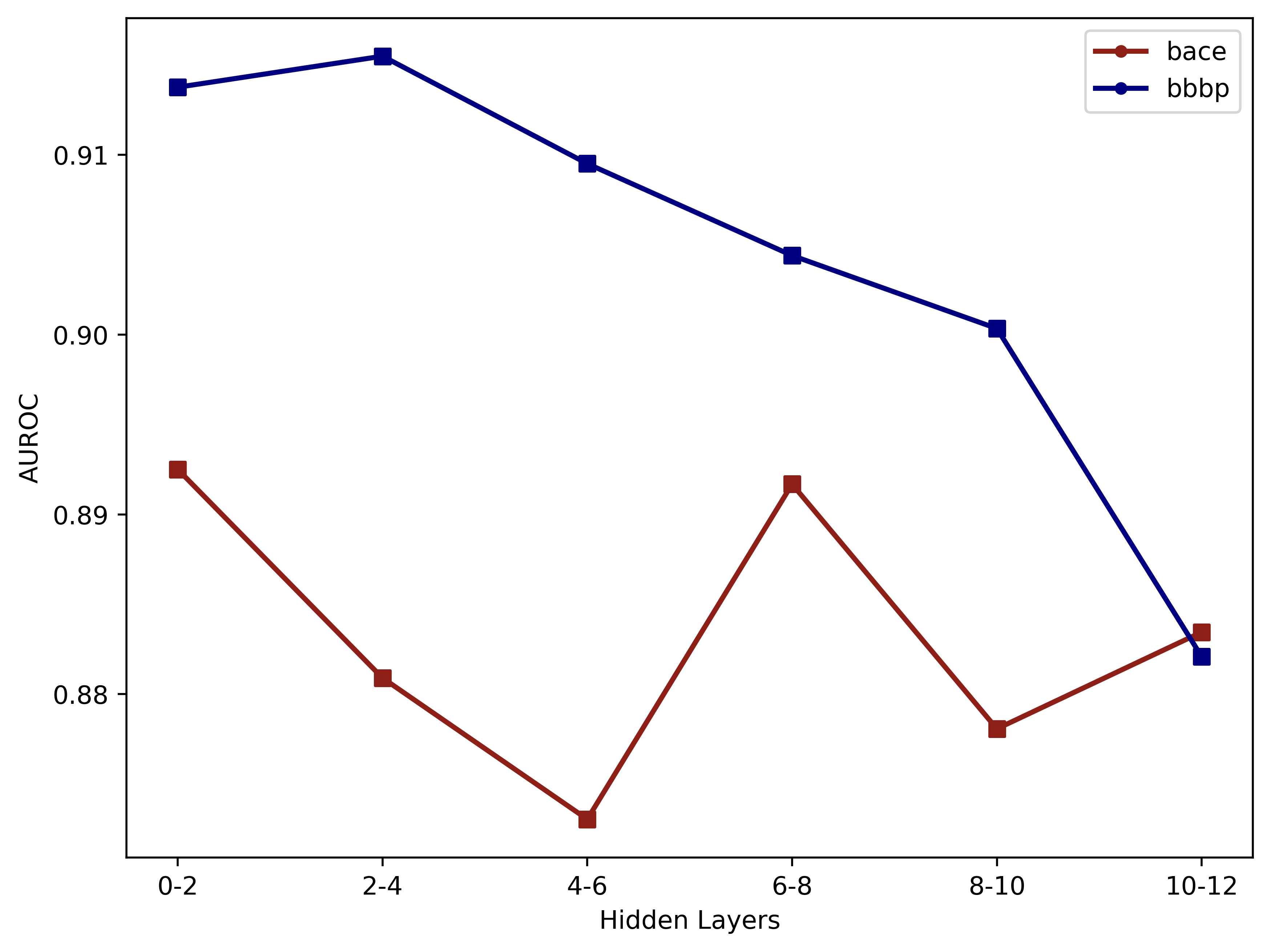}
        \caption{2-layer Steps}
        \label{fig:depth-ablations:a}    
  \end{subfigure}
  \hfill
  \begin{subfigure}{0.48\columnwidth}
      \centering
      \includegraphics[width=\textwidth]{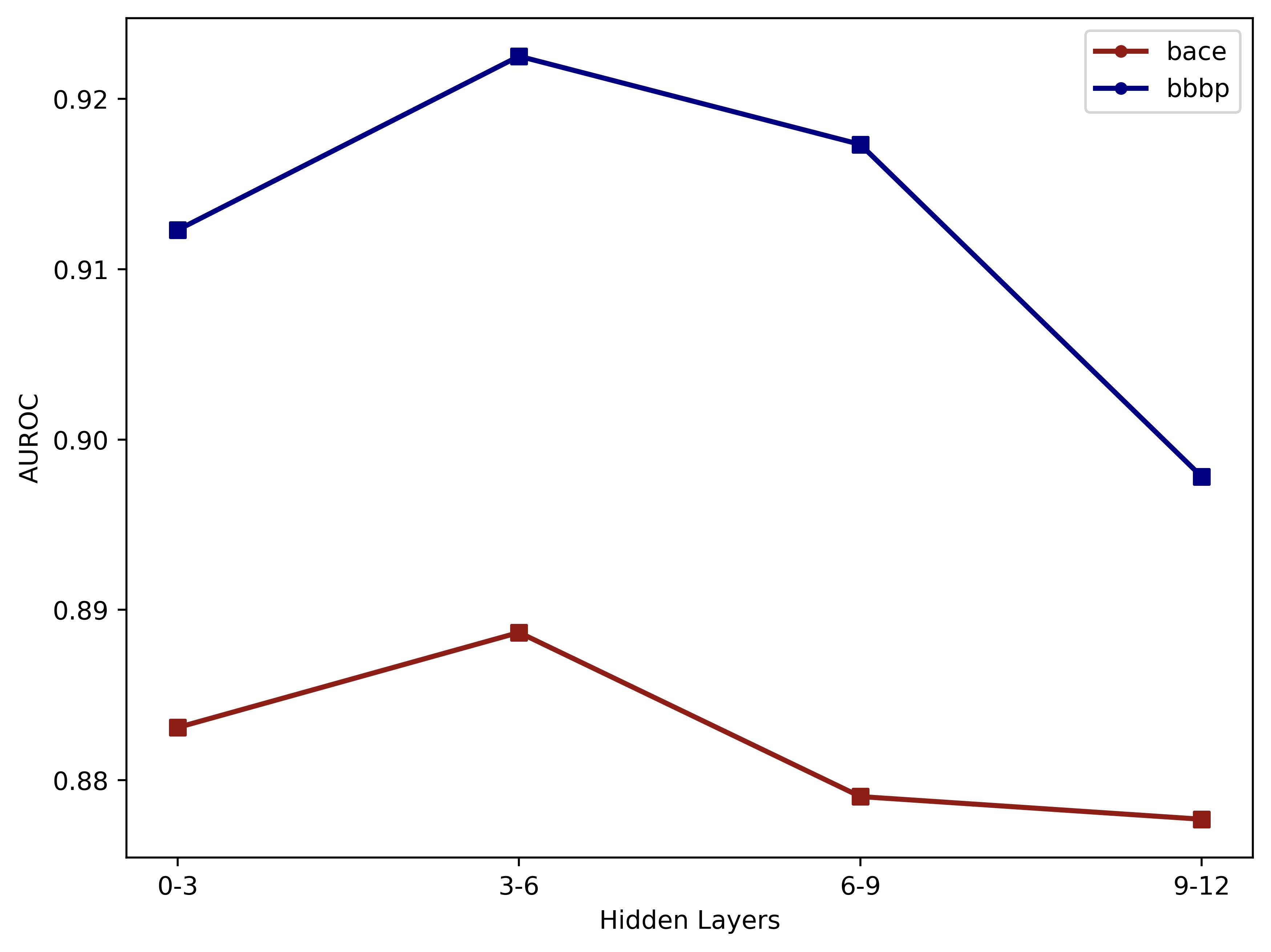}
      \caption{3-layer Steps}
      \label{fig:depth-ablations:b}  
  \end{subfigure}
  \caption{Impact of the depth of injected prompt tokens using Graphormer on BACE and BBBP datasets. Injecting prompts into middle layers works better. Step size does not have much effect.}
\label{fig:depth-ablations}  
\end{figure}

\begin{figure}[ht]
  \begin{subfigure}{0.48\columnwidth}
        \centering
        \includegraphics[width=\textwidth]{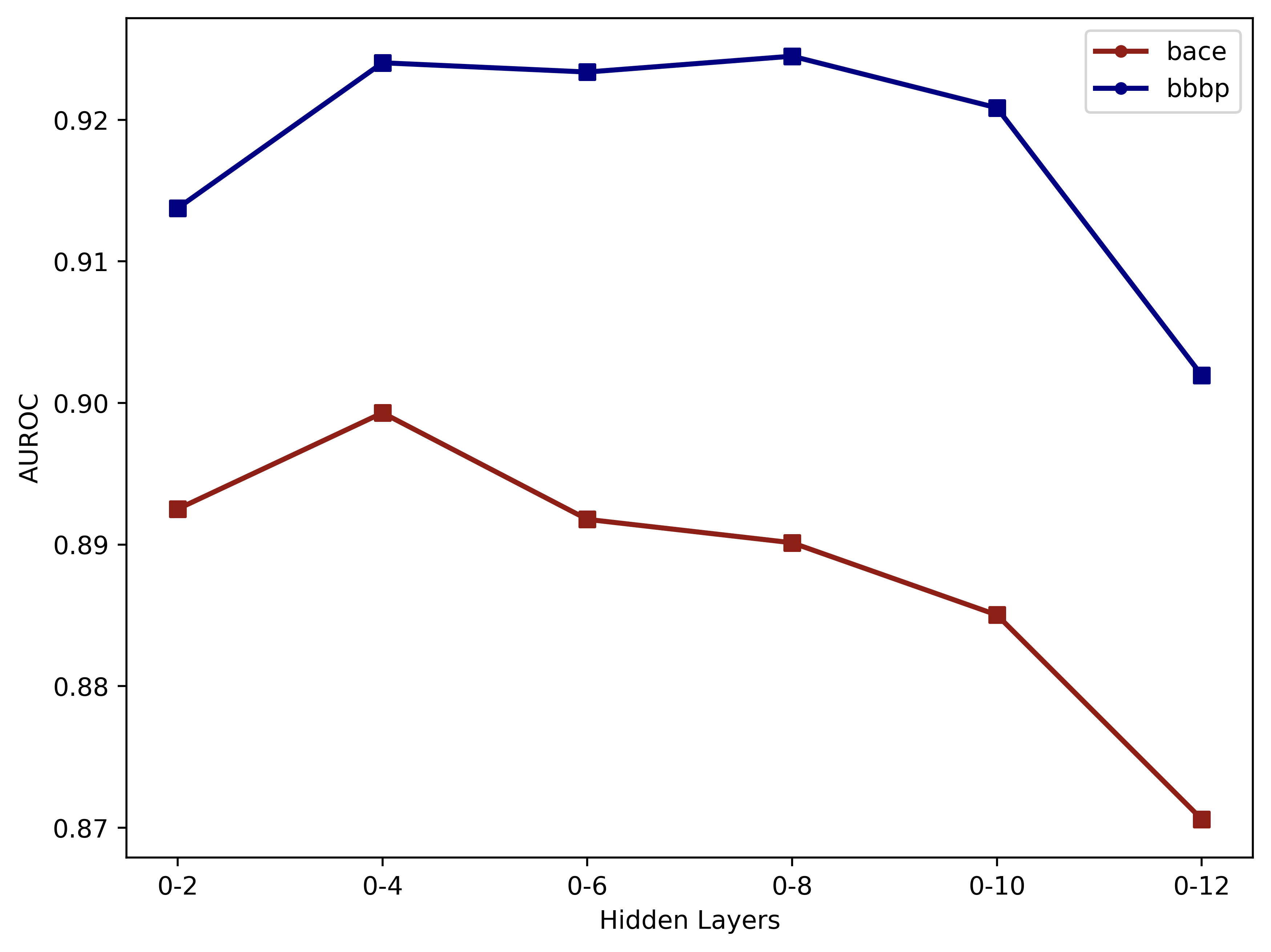}
        \caption{2-layer Steps}
        \label{fig:depth-ablation-asc:a}    
  \end{subfigure}
  \hfill
  \begin{subfigure}{0.48\columnwidth}
      \centering
      \includegraphics[width=\textwidth]{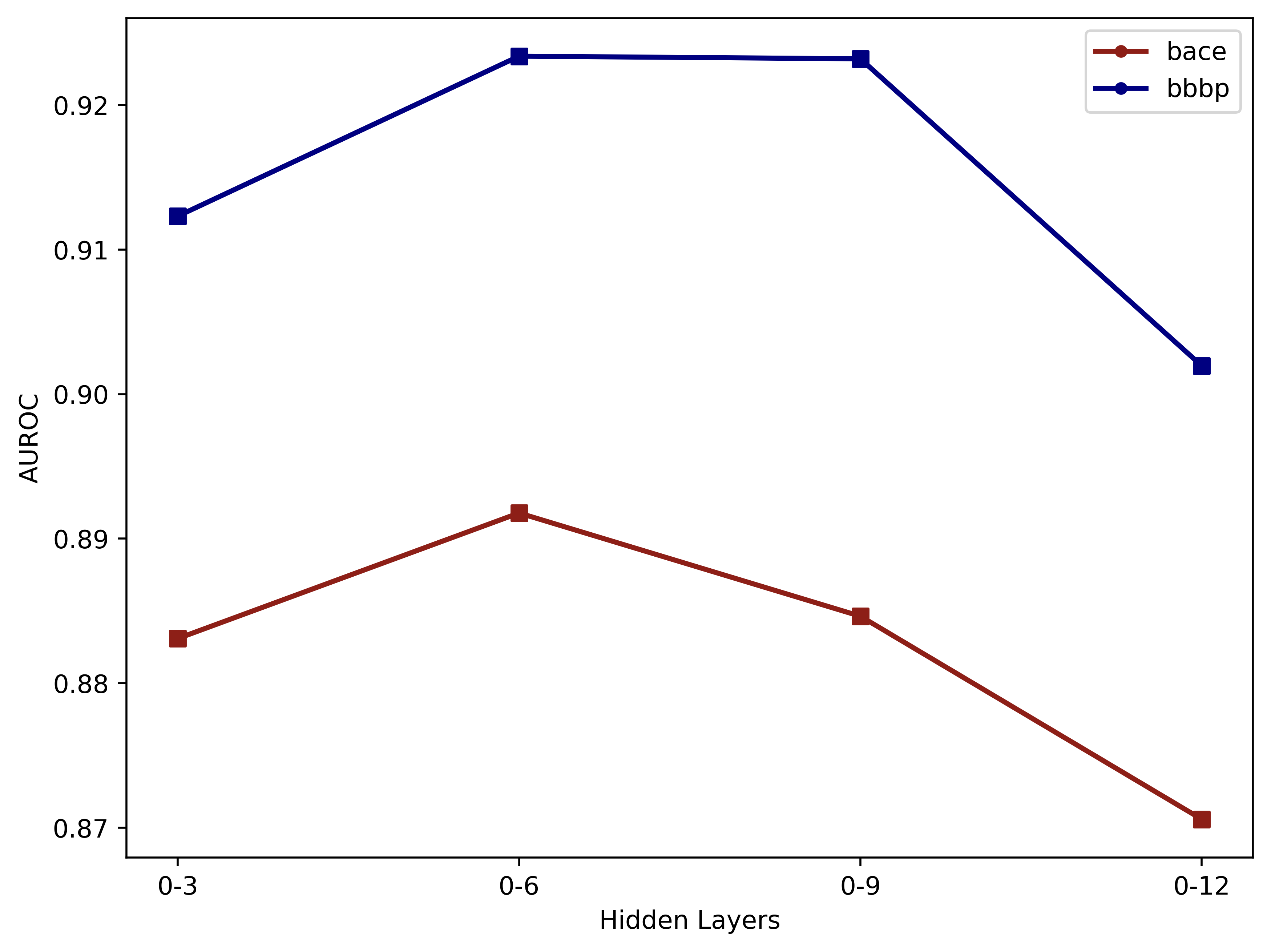}
      \caption{3-layer Steps}
      \label{fig:depth-ablations-asc:b}  
  \end{subfigure}
  \caption{Impact of injecting prompt tokens to initial layers of Graphormer on BACE and BBBP datasets.}
\label{fig:depth-ablations-asc}  
\end{figure}

\begin{figure}[ht]
  \begin{subfigure}{0.48\columnwidth}
        \centering
        \includegraphics[width=\textwidth]{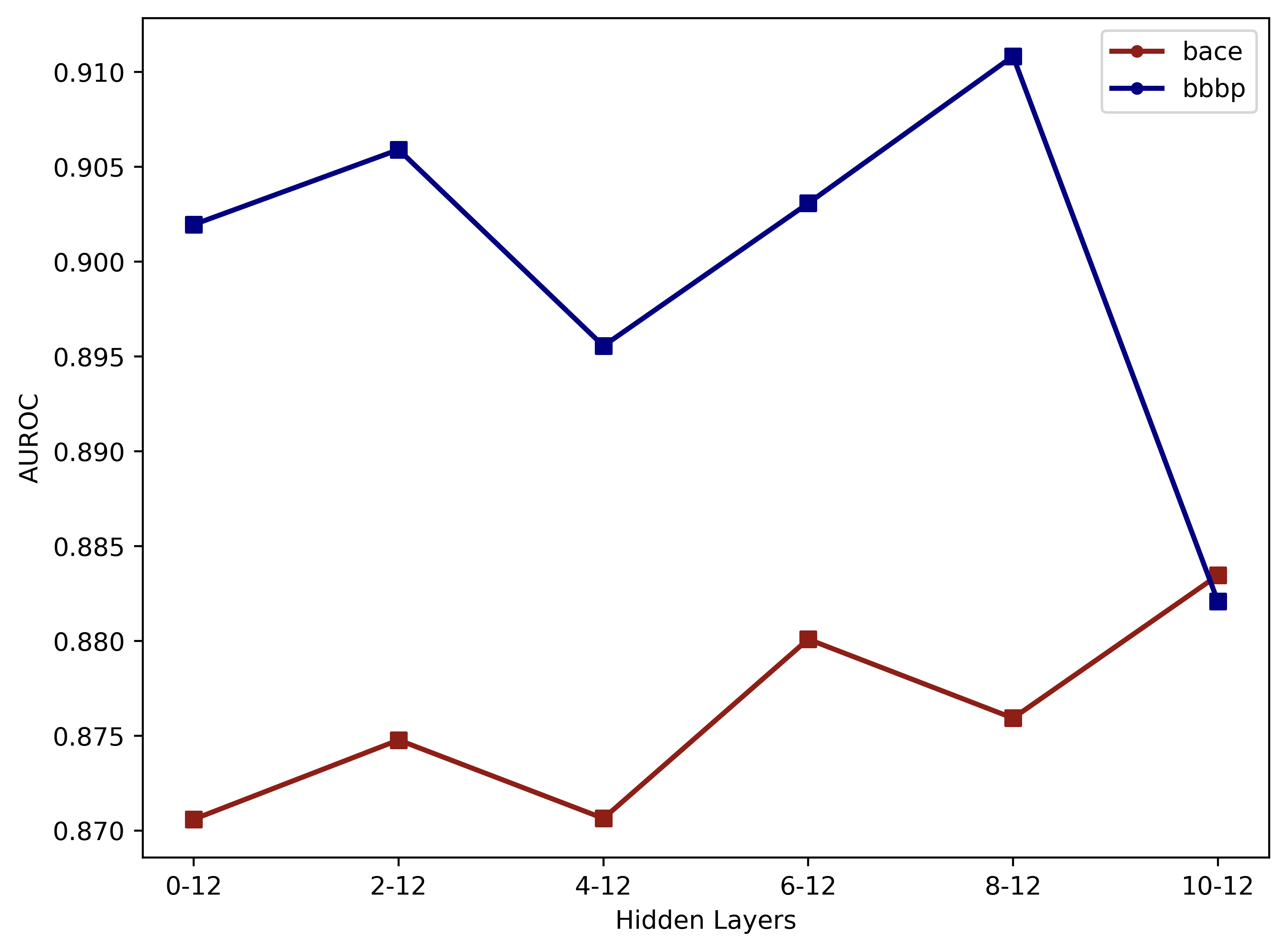}
        \caption{2-layer Steps}
        \label{fig:depth-ablation-desc:a}    
  \end{subfigure}
  \hfill
  \begin{subfigure}{0.48\columnwidth}
      \centering
      \includegraphics[width=\textwidth]{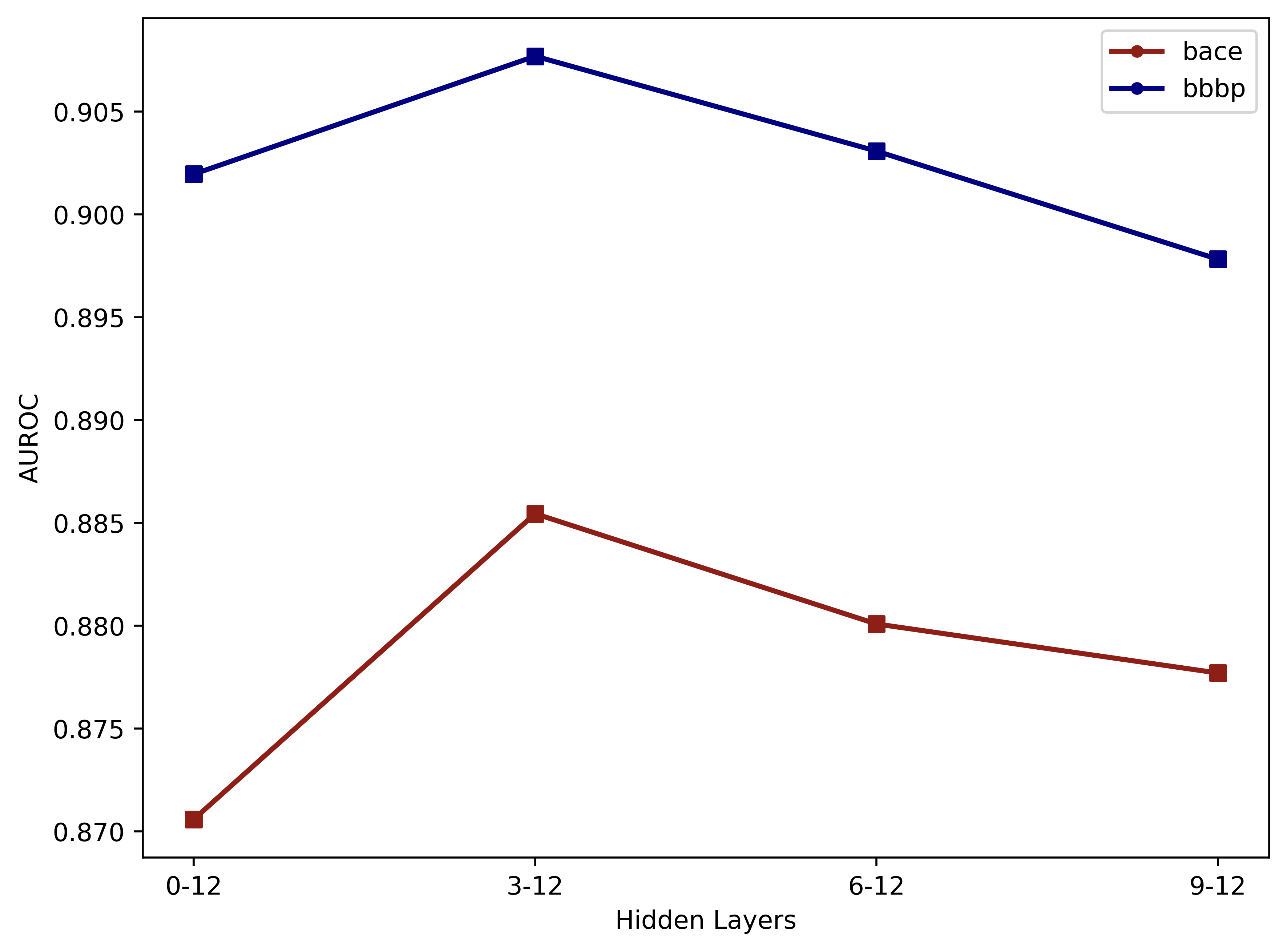}
      \caption{3-layer Steps}
      \label{fig:depth-ablations-desc:b}  
  \end{subfigure}
  \caption{Impact of injecting prompt tokens to final layers of Graphormer on BACE and BBBP datasets.}
\label{fig:depth-ablations-desc}  
\end{figure}

To assess the precise influence of the depth of injected prompt tokens when a certain number of k layers are allocated for prompts, we inject them into different sequences of layers within the graph transformer model. Figure \ref{fig:depth-ablations} illustrates the results. Notably, injecting prompts into middle layers generally produces more favorable outcomes, outperforming other scenarios. We further investigate the effects of injecting prompts into the first or last k layers of the model. The results are shown in Figures \ref{fig:depth-ablations-asc} and \ref{fig:depth-ablations-desc}. Overall, injecting tokens into the middle layers yields better performance, while injecting prompt tokens into the final layers negatively affects the results. Generally the number of layers the prompt tokens are inserted into does not have a significant influence on the results.(For more architectures and datasets, see Appendix).

\subsubsection{Prompt Length}

We also conducted experiments to analyze the effect of prompt length. Figure \ref{fig:ablations-size} illustrates the results, and interestingly, no clear trend is observed for increasing prompt length. However, it is worth noting that, in general, adding more prompt tokens tends to improve performance (For more architectures and datasets, see Appendix).

\begin{figure}[!t]
    \centering
    \includegraphics[width=0.8\columnwidth]{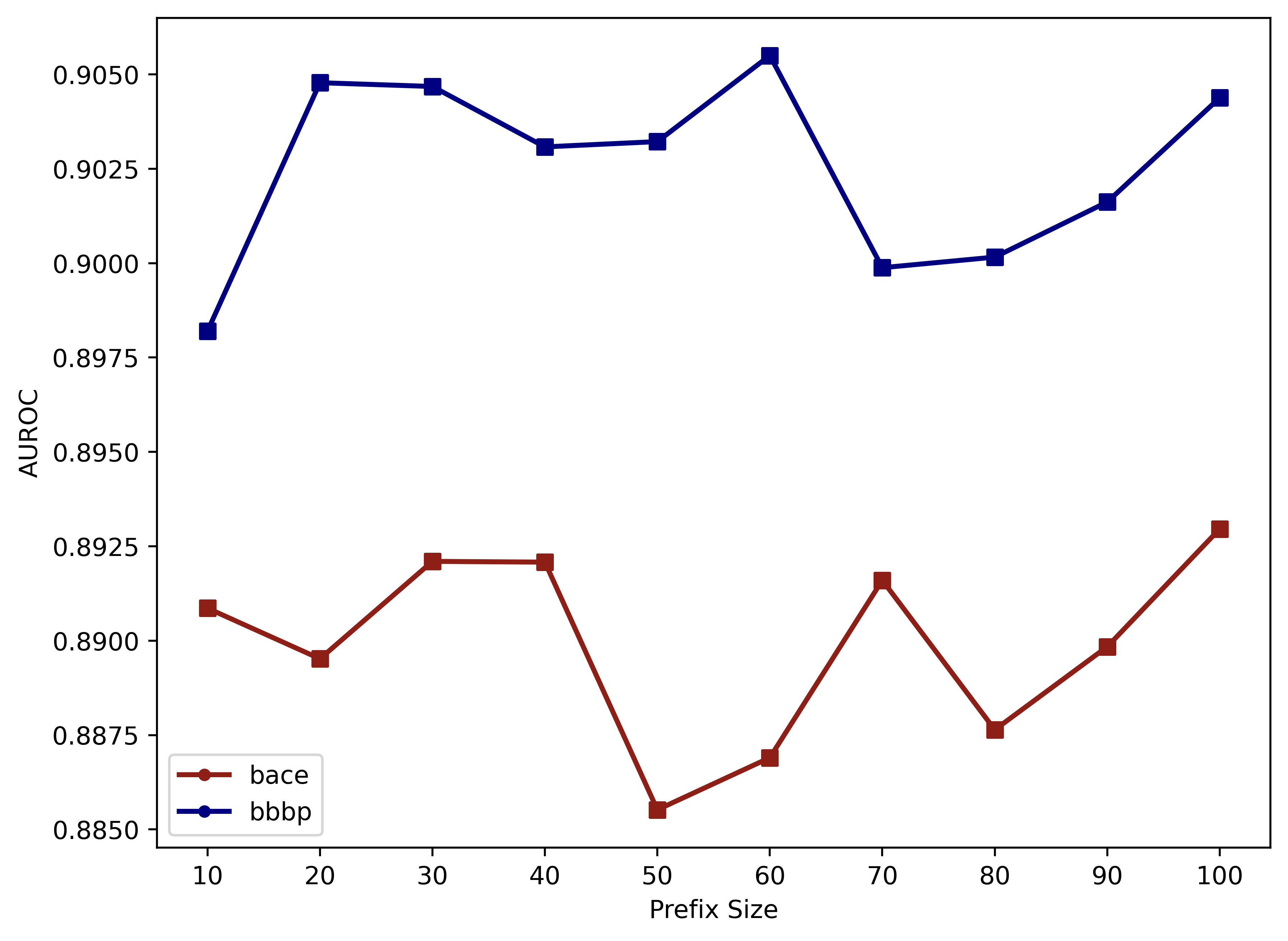}
    \caption{Effect of prompt length using GraphGPS.}
    \label{fig:gps-size-ablations}
\end{figure}

\section{Conclusions}
In this work, we introduced Deep Graph Prompt Tuning (DeepGPT) as an efficient alternative to fine-tuning for graph transformer models in downstream graph based prediction tasks. Our method involves freezing all pre-trained parameters and only tuning task-specific tokens added to the input and the graph transformer layers, reducing computational overhead and storage requirements. Through experiments on various datasets, we showed that DeepGPT achieves comparable or better performance to fine-tuning with significantly fewer task-specific parameters, demonstrating its superiority in resource-intensive applications. This approach addresses the limitations of graph transformers and offers a promising solution for leveraging large pre-trained models in graph-based applications.

\bibliography{references}

\begin{thebibliography}{51}
\providecommand{\natexlab}[1]{#1}

\bibitem[{Alon and Yahav(2021)}]{alon21oversquashing}
Alon, U.; and Yahav, E. 2021.
\newblock On the Bottleneck of Graph Neural Networks and its Practical
  Implications.
\newblock In \emph{9th International Conference on Learning Representations,
  {ICLR} 2021, Virtual Event, Austria, May 3-7, 2021}. OpenReview.net.

\bibitem[{Bresson and Laurent(2017)}]{GatedGCN}
Bresson, X.; and Laurent, T. 2017.
\newblock Residual Gated Graph ConvNets.
\newblock \emph{CoRR}, abs/1711.07553.

\bibitem[{Brown et~al.(2020)Brown, Mann, Ryder, Subbiah, Kaplan, Dhariwal,
  Neelakantan, Shyam, Sastry, Askell, Agarwal, Herbert{-}Voss, Krueger,
  Henighan, Child, Ramesh, Ziegler, Wu, Winter, Hesse, Chen, Sigler, Litwin,
  Gray, Chess, Clark, Berner, McCandlish, Radford, Sutskever, and
  Amodei}]{prompt1}
Brown, T.~B.; Mann, B.; Ryder, N.; Subbiah, M.; Kaplan, J.; Dhariwal, P.;
  Neelakantan, A.; Shyam, P.; Sastry, G.; Askell, A.; Agarwal, S.;
  Herbert{-}Voss, A.; Krueger, G.; Henighan, T.; Child, R.; Ramesh, A.;
  Ziegler, D.~M.; Wu, J.; Winter, C.; Hesse, C.; Chen, M.; Sigler, E.; Litwin,
  M.; Gray, S.; Chess, B.; Clark, J.; Berner, C.; McCandlish, S.; Radford, A.;
  Sutskever, I.; and Amodei, D. 2020.
\newblock Language Models are Few-Shot Learners.
\newblock In Larochelle, H.; Ranzato, M.; Hadsell, R.; Balcan, M.; and Lin, H.,
  eds., \emph{Advances in Neural Information Processing Systems 33: Annual
  Conference on Neural Information Processing Systems 2020, NeurIPS 2020,
  December 6-12, 2020, virtual}.

\bibitem[{Chen et~al.(2020)Chen, Lin, Li, Li, Zhou, and
  Sun}]{chen20oversmoothing}
Chen, D.; Lin, Y.; Li, W.; Li, P.; Zhou, J.; and Sun, X. 2020.
\newblock Measuring and Relieving the Over-Smoothing Problem for Graph Neural
  Networks from the Topological View.
\newblock In \emph{The Thirty-Fourth {AAAI} Conference on Artificial
  Intelligence, {AAAI} 2020, The Thirty-Second Innovative Applications of
  Artificial Intelligence Conference, {IAAI} 2020, The Tenth {AAAI} Symposium
  on Educational Advances in Artificial Intelligence, {EAAI} 2020, New York,
  NY, USA, February 7-12, 2020}, 3438--3445. {AAAI} Press.

\bibitem[{Chen, O'Bray, and Borgwardt(2022)}]{chen2022sat}
Chen, D.; O'Bray, L.; and Borgwardt, K.~M. 2022.
\newblock Structure-Aware Transformer for Graph Representation Learning.
\newblock In Chaudhuri, K.; Jegelka, S.; Song, L.; Szepesv{\'{a}}ri, C.; Niu,
  G.; and Sabato, S., eds., \emph{International Conference on Machine Learning,
  {ICML} 2022, 17-23 July 2022, Baltimore, Maryland, {USA}}, volume 162 of
  \emph{Proceedings of Machine Learning Research}, 3469--3489. {PMLR}.

\bibitem[{Chithrananda, Grand, and Ramsundar(2020)}]{chemberta}
Chithrananda, S.; Grand, G.; and Ramsundar, B. 2020.
\newblock ChemBERTa: Large-Scale Self-Supervised Pretraining for Molecular
  Property Prediction.
\newblock \emph{CoRR}, abs/2010.09885.

\bibitem[{Choromanski et~al.(2021)Choromanski, Likhosherstov, Dohan, Song,
  Gane, Sarl{\'{o}}s, Hawkins, Davis, Mohiuddin, Kaiser, Belanger, Colwell, and
  Weller}]{performer}
Choromanski, K.~M.; Likhosherstov, V.; Dohan, D.; Song, X.; Gane, A.;
  Sarl{\'{o}}s, T.; Hawkins, P.; Davis, J.~Q.; Mohiuddin, A.; Kaiser, L.;
  Belanger, D.~B.; Colwell, L.~J.; and Weller, A. 2021.
\newblock Rethinking Attention with Performers.
\newblock In \emph{9th International Conference on Learning Representations,
  {ICLR} 2021, Virtual Event, Austria, May 3-7, 2021}. OpenReview.net.

\bibitem[{Corso et~al.(2020)Corso, Cavalleri, Beaini, Li{\`{o}}, and
  Velickovic}]{PNA}
Corso, G.; Cavalleri, L.; Beaini, D.; Li{\`{o}}, P.; and Velickovic, P. 2020.
\newblock Principal Neighbourhood Aggregation for Graph Nets.
\newblock \emph{CoRR}, abs/2004.05718.

\bibitem[{Devlin et~al.(2019)Devlin, Chang, Lee, and Toutanova}]{bert}
Devlin, J.; Chang, M.; Lee, K.; and Toutanova, K. 2019.
\newblock {BERT:} Pre-training of Deep Bidirectional Transformers for Language
  Understanding.
\newblock In Burstein, J.; Doran, C.; and Solorio, T., eds., \emph{Proceedings
  of the 2019 Conference of the North American Chapter of the Association for
  Computational Linguistics: Human Language Technologies, {NAACL-HLT} 2019,
  Minneapolis, MN, USA, June 2-7, 2019, Volume 1 (Long and Short Papers)},
  4171--4186. Association for Computational Linguistics.

\bibitem[{Dosovitskiy et~al.(2021)Dosovitskiy, Beyer, Kolesnikov, Weissenborn,
  Zhai, Unterthiner, Dehghani, Minderer, Heigold, Gelly, Uszkoreit, and
  Houlsby}]{dosovitskiy2021vit}
Dosovitskiy, A.; Beyer, L.; Kolesnikov, A.; Weissenborn, D.; Zhai, X.;
  Unterthiner, T.; Dehghani, M.; Minderer, M.; Heigold, G.; Gelly, S.;
  Uszkoreit, J.; and Houlsby, N. 2021.
\newblock An Image is Worth 16x16 Words: Transformers for Image Recognition at
  Scale.
\newblock In \emph{9th International Conference on Learning Representations,
  {ICLR} 2021, Virtual Event, Austria, May 3-7, 2021}. OpenReview.net.

\bibitem[{Dwivedi and Bresson(2020)}]{dwivedi2020gt}
Dwivedi, V.~P.; and Bresson, X. 2020.
\newblock A Generalization of Transformer Networks to Graphs.
\newblock \emph{CoRR}, abs/2012.09699.

\bibitem[{Dwivedi et~al.(2022)Dwivedi, Luu, Laurent, Bengio, and
  Bresson}]{dwivedi2022rwpe}
Dwivedi, V.~P.; Luu, A.~T.; Laurent, T.; Bengio, Y.; and Bresson, X. 2022.
\newblock Graph Neural Networks with Learnable Structural and Positional
  Representations.
\newblock In \emph{International Conference on Learning Representations}.

\bibitem[{Fang et~al.(2022)Fang, Zhang, Yang, and Wang}]{fang22graphprompt}
Fang, T.; Zhang, Y.; Yang, Y.; and Wang, C. 2022.
\newblock Prompt Tuning for Graph Neural Networks.
\newblock \emph{CoRR}, abs/2209.15240.

\bibitem[{Feng et~al.(2022)Feng, Chen, Li, Sarkar, and
  Zhang}]{feng22mpgexpress}
Feng, J.; Chen, Y.; Li, F.; Sarkar, A.; and Zhang, M. 2022.
\newblock How Powerful are K-hop Message Passing Graph Neural Networks.
\newblock In \emph{NeurIPS}.

\bibitem[{Gao, Wang, and Ji(2018)}]{gao2018nodelevel}
Gao, H.; Wang, Z.; and Ji, S. 2018.
\newblock Large-Scale Learnable Graph Convolutional Networks.
\newblock In \emph{Proceedings of the 24th ACM SIGKDD International Conference
  on Knowledge Discovery \& Data Mining}, KDD '18, 1416–1424. New York, NY,
  USA: Association for Computing Machinery.
\newblock ISBN 9781450355520.

\bibitem[{Gao, Fisch, and Chen(2021)}]{prompt2}
Gao, T.; Fisch, A.; and Chen, D. 2021.
\newblock Making Pre-trained Language Models Better Few-shot Learners.
\newblock In Zong, C.; Xia, F.; Li, W.; and Navigli, R., eds.,
  \emph{Proceedings of the 59th Annual Meeting of the Association for
  Computational Linguistics and the 11th International Joint Conference on
  Natural Language Processing, {ACL/IJCNLP} 2021, (Volume 1: Long Papers),
  Virtual Event, August 1-6, 2021}, 3816--3830. Association for Computational
  Linguistics.

\bibitem[{Gaulton et~al.(2012)Gaulton, Bellis, Bento, Chambers, Davies, Hersey,
  Light, McGlinchey, Michalovich, Al{-}Lazikani, and Overington}]{chembl}
Gaulton, A.; Bellis, L.~J.; Bento, A.~P.; Chambers, J.; Davies, M.; Hersey, A.;
  Light, Y.; McGlinchey, S.; Michalovich, D.; Al{-}Lazikani, B.; and
  Overington, J.~P. 2012.
\newblock ChEMBL: a large-scale bioactivity database for drug discovery.
\newblock \emph{Nucleic Acids Res.}, 40(Database-Issue): 1100--1107.

\bibitem[{Gilmer et~al.(2017)Gilmer, Schoenholz, Riley, Vinyals, and
  Dahl}]{gilmer2017messagepassing}
Gilmer, J.; Schoenholz, S.~S.; Riley, P.~F.; Vinyals, O.; and Dahl, G.~E. 2017.
\newblock Neural Message Passing for Quantum Chemistry.
\newblock In \emph{Proceedings of the 34th International Conference on Machine
  Learning - Volume 70}, ICML'17, 1263–1272. JMLR.org.

\bibitem[{Hao et~al.(2020)Hao, Lu, Huang, Wang, Hu, Liu, Chen, and
  Lee}]{hao20graphlevel}
Hao, Z.; Lu, C.; Huang, Z.; Wang, H.; Hu, Z.; Liu, Q.; Chen, E.; and Lee, C.
  2020.
\newblock ASGN: An Active Semi-Supervised Graph Neural Network for Molecular
  Property Prediction.
\newblock In \emph{Proceedings of the 26th ACM SIGKDD International Conference
  on Knowledge Discovery \& Data Mining}, KDD '20, 731–752. New York, NY,
  USA: Association for Computing Machinery.
\newblock ISBN 9781450379984.

\bibitem[{Howard and Ruder(2018)}]{fine-tuning}
Howard, J.; and Ruder, S. 2018.
\newblock Universal Language Model Fine-tuning for Text Classification.
\newblock In Gurevych, I.; and Miyao, Y., eds., \emph{Proceedings of the 56th
  Annual Meeting of the Association for Computational Linguistics, {ACL} 2018,
  Melbourne, Australia, July 15-20, 2018, Volume 1: Long Papers}, 328--339.
  Association for Computational Linguistics.

\bibitem[{Hu et~al.(2021)Hu, Fey, Ren, Nakata, Dong, and
  Leskovec}]{hu2021ogblsc}
Hu, W.; Fey, M.; Ren, H.; Nakata, M.; Dong, Y.; and Leskovec, J. 2021.
\newblock Ogb-lsc: A large-scale challenge for machine learning on graphs.
\newblock \emph{arXiv preprint arXiv:2103.09430}.

\bibitem[{Hu et~al.(2020{\natexlab{a}})Hu, Fey, Zitnik, Dong, Ren, Liu,
  Catasta, and Leskovec}]{hu2020ogb}
Hu, W.; Fey, M.; Zitnik, M.; Dong, Y.; Ren, H.; Liu, B.; Catasta, M.; and
  Leskovec, J. 2020{\natexlab{a}}.
\newblock Open graph benchmark: Datasets for machine learning on graphs.
\newblock \emph{Advances in neural information processing systems}, 33:
  22118--22133.

\bibitem[{Hu et~al.(2020{\natexlab{b}})Hu, Liu, Gomes, Zitnik, Liang, Pande,
  and Leskovec}]{GINE}
Hu, W.; Liu, B.; Gomes, J.; Zitnik, M.; Liang, P.; Pande, V.~S.; and Leskovec,
  J. 2020{\natexlab{b}}.
\newblock Strategies for Pre-training Graph Neural Networks.
\newblock In \emph{8th International Conference on Learning Representations,
  {ICLR} 2020, Addis Ababa, Ethiopia, April 26-30, 2020}. OpenReview.net.

\bibitem[{Hu et~al.(2020{\natexlab{c}})Hu, Dong, Wang, Chang, and
  Sun}]{hu2020generativepretraining}
Hu, Z.; Dong, Y.; Wang, K.; Chang, K.-W.; and Sun, Y. 2020{\natexlab{c}}.
\newblock GPT-GNN: Generative Pre-Training of Graph Neural Networks.
\newblock In \emph{Proceedings of the 26th ACM SIGKDD International Conference
  on Knowledge Discovery \& Data Mining}, KDD '20, 1857–1867. New York, NY,
  USA: Association for Computing Machinery.
\newblock ISBN 9781450379984.

\bibitem[{Kreuzer et~al.(2021)Kreuzer, Beaini, Hamilton, L{\'{e}}tourneau, and
  Tossou}]{kruezer2021san}
Kreuzer, D.; Beaini, D.; Hamilton, W.~L.; L{\'{e}}tourneau, V.; and Tossou, P.
  2021.
\newblock Rethinking Graph Transformers with Spectral Attention.
\newblock In Ranzato, M.; Beygelzimer, A.; Dauphin, Y.~N.; Liang, P.; and
  Vaughan, J.~W., eds., \emph{Advances in Neural Information Processing Systems
  34: Annual Conference on Neural Information Processing Systems 2021, NeurIPS
  2021, December 6-14, 2021, virtual}, 21618--21629.

\bibitem[{Lester, Al{-}Rfou, and Constant(2021)}]{lester21promptnlp1}
Lester, B.; Al{-}Rfou, R.; and Constant, N. 2021.
\newblock The Power of Scale for Parameter-Efficient Prompt Tuning.
\newblock In Moens, M.; Huang, X.; Specia, L.; and Yih, S.~W., eds.,
  \emph{Proceedings of the 2021 Conference on Empirical Methods in Natural
  Language Processing, {EMNLP} 2021, Virtual Event / Punta Cana, Dominican
  Republic, 7-11 November, 2021}, 3045--3059. Association for Computational
  Linguistics.

\bibitem[{Li, Zhao, and Zeng(2022)}]{han2022kpgt}
Li, H.; Zhao, D.; and Zeng, J. 2022.
\newblock {KPGT:} Knowledge-Guided Pre-training of Graph Transformer for
  Molecular Property Prediction.
\newblock In Zhang, A.; and Rangwala, H., eds., \emph{{KDD} '22: The 28th {ACM}
  {SIGKDD} Conference on Knowledge Discovery and Data Mining, Washington, DC,
  USA, August 14 - 18, 2022}, 857--867. {ACM}.

\bibitem[{Li et~al.(2021)Li, Zhou, Xu, Huang, Wang, Xiong, Huang, Dou, and
  Xiong}]{li21edgelevel}
Li, S.; Zhou, J.; Xu, T.; Huang, L.; Wang, F.; Xiong, H.; Huang, W.; Dou, D.;
  and Xiong, H. 2021.
\newblock Structure-Aware Interactive Graph Neural Networks for the Prediction
  of Protein-Ligand Binding Affinity.
\newblock In \emph{Proceedings of the 27th ACM SIGKDD Conference on Knowledge
  Discovery \& Data Mining}, KDD '21, 975–985. New York, NY, USA: Association
  for Computing Machinery.
\newblock ISBN 9781450383325.

\bibitem[{Li and Liang(2021)}]{lia21prefixtuning}
Li, X.~L.; and Liang, P. 2021.
\newblock Prefix-Tuning: Optimizing Continuous Prompts for Generation.
\newblock In Zong, C.; Xia, F.; Li, W.; and Navigli, R., eds.,
  \emph{Proceedings of the 59th Annual Meeting of the Association for
  Computational Linguistics and the 11th International Joint Conference on
  Natural Language Processing, {ACL/IJCNLP} 2021, (Volume 1: Long Papers),
  Virtual Event, August 1-6, 2021}, 4582--4597. Association for Computational
  Linguistics.

\bibitem[{Liu et~al.(2021{\natexlab{a}})Liu, Ji, Fu, Du, Yang, and
  Tang}]{p-tuning-v2}
Liu, X.; Ji, K.; Fu, Y.; Du, Z.; Yang, Z.; and Tang, J. 2021{\natexlab{a}}.
\newblock P-Tuning v2: Prompt Tuning Can Be Comparable to Fine-tuning
  Universally Across Scales and Tasks.
\newblock \emph{CoRR}, abs/2110.07602.

\bibitem[{Liu et~al.(2021{\natexlab{b}})Liu, Zheng, Du, Ding, Qian, Yang, and
  Tang}]{promptnlp2}
Liu, X.; Zheng, Y.; Du, Z.; Ding, M.; Qian, Y.; Yang, Z.; and Tang, J.
  2021{\natexlab{b}}.
\newblock {GPT} Understands, Too.
\newblock \emph{CoRR}, abs/2103.10385.

\bibitem[{Liu et~al.(2019)Liu, Ott, Goyal, Du, Joshi, Chen, Levy, Lewis,
  Zettlemoyer, and Stoyanov}]{roberta}
Liu, Y.; Ott, M.; Goyal, N.; Du, J.; Joshi, M.; Chen, D.; Levy, O.; Lewis, M.;
  Zettlemoyer, L.; and Stoyanov, V. 2019.
\newblock RoBERTa: {A} Robustly Optimized {BERT} Pretraining Approach.
\newblock \emph{CoRR}, abs/1907.11692.

\bibitem[{Loshchilov and Hutter(2019)}]{adamw}
Loshchilov, I.; and Hutter, F. 2019.
\newblock Decoupled Weight Decay Regularization.
\newblock In \emph{7th International Conference on Learning Representations,
  {ICLR} 2019, New Orleans, LA, USA, May 6-9, 2019}. OpenReview.net.

\bibitem[{Minaee et~al.(2021)Minaee, Kalchbrenner, Cambria, Nikzad, Chenaghlu,
  and Gao}]{text-classification}
Minaee, S.; Kalchbrenner, N.; Cambria, E.; Nikzad, N.; Chenaghlu, M.; and Gao,
  J. 2021.
\newblock Deep learning--based text classification: a comprehensive review.
\newblock \emph{ACM computing surveys (CSUR)}, 54(3): 1--40.

\bibitem[{Podlewska and Kafel(2018)}]{metstab}
Podlewska, S.; and Kafel, R. 2018.
\newblock MetStabOn—online platform for metabolic stability predictions.
\newblock \emph{International journal of molecular sciences}, 19(4): 1040.

\bibitem[{Qiu et~al.(2020)Qiu, Chen, Dong, Zhang, Yang, Ding, Wang, and
  Tang}]{qui2020contrastive-1}
Qiu, J.; Chen, Q.; Dong, Y.; Zhang, J.; Yang, H.; Ding, M.; Wang, K.; and Tang,
  J. 2020.
\newblock GCC: Graph Contrastive Coding for Graph Neural Network Pre-Training.
\newblock In \emph{Proceedings of the 26th ACM SIGKDD International Conference
  on Knowledge Discovery \& Data Mining}, KDD '20, 1150–1160. New York, NY,
  USA: Association for Computing Machinery.
\newblock ISBN 9781450379984.

\bibitem[{Radford et~al.(2019)Radford, Wu, Child, Luan, Amodei, Sutskever
  et~al.}]{radford2019language}
Radford, A.; Wu, J.; Child, R.; Luan, D.; Amodei, D.; Sutskever, I.; et~al.
  2019.
\newblock Language models are unsupervised multitask learners.
\newblock \emph{OpenAI blog}, 1(8): 9.

\bibitem[{Ramp{\'{a}}sek et~al.(2022)Ramp{\'{a}}sek, Galkin, Dwivedi, Luu,
  Wolf, and Beaini}]{ladislav2022graphgps}
Ramp{\'{a}}sek, L.; Galkin, M.; Dwivedi, V.~P.; Luu, A.~T.; Wolf, G.; and
  Beaini, D. 2022.
\newblock Recipe for a General, Powerful, Scalable Graph Transformer.
\newblock In \emph{NeurIPS}.

\bibitem[{Rong et~al.(2020)Rong, Bian, Xu, Xie, Wei, Huang, and
  Huang}]{selfsupervised-gt}
Rong, Y.; Bian, Y.; Xu, T.; Xie, W.; Wei, Y.; Huang, W.; and Huang, J. 2020.
\newblock Self-Supervised Graph Transformer on Large-Scale Molecular Data.
\newblock In Larochelle, H.; Ranzato, M.; Hadsell, R.; Balcan, M.; and Lin, H.,
  eds., \emph{Advances in Neural Information Processing Systems 33: Annual
  Conference on Neural Information Processing Systems 2020, NeurIPS 2020,
  December 6-12, 2020, virtual}.

\bibitem[{Shin et~al.(2020)Shin, Razeghi, IV, Wallace, and Singh}]{autoprompt}
Shin, T.; Razeghi, Y.; IV, R. L.~L.; Wallace, E.; and Singh, S. 2020.
\newblock AutoPrompt: Eliciting Knowledge from Language Models with
  Automatically Generated Prompts.
\newblock In Webber, B.; Cohn, T.; He, Y.; and Liu, Y., eds., \emph{Proceedings
  of the 2020 Conference on Empirical Methods in Natural Language Processing,
  {EMNLP} 2020, Online, November 16-20, 2020}, 4222--4235. Association for
  Computational Linguistics.

\bibitem[{Subramonian(2021)}]{subramonian2021contrastive2}
Subramonian, A. 2021.
\newblock Motif-driven contrastive learning of graph representations.
\newblock In \emph{Proceedings of the AAAI Conference on Artificial
  Intelligence}, volume~35, 15980--15981.

\bibitem[{Sun et~al.(2022)Sun, Zhou, He, Wang, and Wang}]{sun22gppt}
Sun, M.; Zhou, K.; He, X.; Wang, Y.; and Wang, X. 2022.
\newblock GPPT: Graph Pre-Training and Prompt Tuning to Generalize Graph Neural
  Networks.
\newblock In \emph{Proceedings of the 28th ACM SIGKDD Conference on Knowledge
  Discovery and Data Mining}, KDD '22, 1717–1727. New York, NY, USA:
  Association for Computing Machinery.
\newblock ISBN 9781450393850.

\bibitem[{Vaswani et~al.(2017)Vaswani, Shazeer, Parmar, Uszkoreit, Jones,
  Gomez, Kaiser, and Polosukhin}]{vaswani2017attention}
Vaswani, A.; Shazeer, N.; Parmar, N.; Uszkoreit, J.; Jones, L.; Gomez, A.~N.;
  Kaiser, L.; and Polosukhin, I. 2017.
\newblock Attention is All you Need.
\newblock In Guyon, I.; von Luxburg, U.; Bengio, S.; Wallach, H.~M.; Fergus,
  R.; Vishwanathan, S. V.~N.; and Garnett, R., eds., \emph{Advances in Neural
  Information Processing Systems 30: Annual Conference on Neural Information
  Processing Systems 2017, December 4-9, 2017, Long Beach, CA, {USA}},
  5998--6008.

\bibitem[{Velickovic et~al.(2019)Velickovic, Fedus, Hamilton, Li{\`{o}},
  Bengio, and Hjelm}]{DGI}
Velickovic, P.; Fedus, W.; Hamilton, W.~L.; Li{\`{o}}, P.; Bengio, Y.; and
  Hjelm, R.~D. 2019.
\newblock Deep Graph Infomax.
\newblock In \emph{7th International Conference on Learning Representations,
  {ICLR} 2019, New Orleans, LA, USA, May 6-9, 2019}. OpenReview.net.

\bibitem[{Wang and Zhang(2022)}]{wang22spectralexpress}
Wang, X.; and Zhang, M. 2022.
\newblock How Powerful are Spectral Graph Neural Networks.
\newblock In Chaudhuri, K.; Jegelka, S.; Song, L.; Szepesv{\'{a}}ri, C.; Niu,
  G.; and Sabato, S., eds., \emph{International Conference on Machine Learning,
  {ICML} 2022, 17-23 July 2022, Baltimore, Maryland, {USA}}, volume 162 of
  \emph{Proceedings of Machine Learning Research}, 23341--23362. {PMLR}.

\bibitem[{Wu et~al.(2018)Wu, Ramsundar, Feinberg, Gomes, Geniesse, Pappu,
  Leswing, and Pande}]{wu2018moleculenet}
Wu, Z.; Ramsundar, B.; Feinberg, E.~N.; Gomes, J.; Geniesse, C.; Pappu, A.~S.;
  Leswing, K.; and Pande, V. 2018.
\newblock MoleculeNet: a benchmark for molecular machine learning.
\newblock \emph{Chemical science}, 9(2): 513--530.

\bibitem[{Ying et~al.(2021)Ying, Cai, Luo, Zheng, Ke, He, Shen, and
  Liu}]{ying2021graphormer}
Ying, C.; Cai, T.; Luo, S.; Zheng, S.; Ke, G.; He, D.; Shen, Y.; and Liu, T.
  2021.
\newblock Do Transformers Really Perform Badly for Graph Representation?
\newblock In Ranzato, M.; Beygelzimer, A.; Dauphin, Y.~N.; Liang, P.; and
  Vaughan, J.~W., eds., \emph{Advances in Neural Information Processing Systems
  34: Annual Conference on Neural Information Processing Systems 2021, NeurIPS
  2021, December 6-14, 2021, virtual}, 28877--28888.

\bibitem[{You et~al.(2021)You, Chen, Shen, and Wang}]{GCL2}
You, Y.; Chen, T.; Shen, Y.; and Wang, Z. 2021.
\newblock Graph Contrastive Learning Automated.
\newblock In Meila, M.; and Zhang, T., eds., \emph{Proceedings of the 38th
  International Conference on Machine Learning, {ICML} 2021, 18-24 July 2021,
  Virtual Event}, volume 139 of \emph{Proceedings of Machine Learning
  Research}, 12121--12132. {PMLR}.

\bibitem[{You et~al.(2020)You, Chen, Sui, Chen, Wang, and Shen}]{GraphCL}
You, Y.; Chen, T.; Sui, Y.; Chen, T.; Wang, Z.; and Shen, Y. 2020.
\newblock Graph Contrastive Learning with Augmentations.
\newblock In Larochelle, H.; Ranzato, M.; Hadsell, R.; Balcan, M.; and Lin, H.,
  eds., \emph{Advances in Neural Information Processing Systems 33: Annual
  Conference on Neural Information Processing Systems 2020, NeurIPS 2020,
  December 6-12, 2020, virtual}.

\bibitem[{Zhang, Wang, and Liu(2018)}]{sentiment-analysis}
Zhang, L.; Wang, S.; and Liu, B. 2018.
\newblock Deep learning for sentiment analysis: A survey.
\newblock \emph{Wiley Interdisciplinary Reviews: Data Mining and Knowledge
  Discovery}, 8(4): e1253.

\bibitem[{Zhang et~al.(2021)Zhang, Liu, Wang, Lu, and Lee}]{MGSSL}
Zhang, Z.; Liu, Q.; Wang, H.; Lu, C.; and Lee, C. 2021.
\newblock Motif-based Graph Self-Supervised Learning for Molecular Property
  Prediction.
\newblock In Ranzato, M.; Beygelzimer, A.; Dauphin, Y.~N.; Liang, P.; and
  Vaughan, J.~W., eds., \emph{Advances in Neural Information Processing Systems
  34: Annual Conference on Neural Information Processing Systems 2021, NeurIPS
  2021, December 6-14, 2021, virtual}, 15870--15882.

\end{thebibliography}

\clearpage
\appendix

\section{Experimental Details}

\subsection{Datasets Description}
\label{sec-dataset-desc}

\begin{table}[htbp]
\centering
\caption{Overview of graph-level prediction datasets used in this study.}
\label{tab:datasets}
\resizebox{\columnwidth}{!}{%
\begin{tabular}{@{}lccc@{}}
\toprule
\textbf{Dataset}         & \textbf{\#Graphs} & \textbf{Prediction Task} & \textbf{Metric} \\ \midrule
BACE\cite{wu2018moleculenet}           & 1513    & binary classification & AUROC\\
BBBP\cite{wu2018moleculenet}            & 2039    & binary classification & AUROC\\ 
ClinTox\cite{wu2018moleculenet}         & 1478    & 2-task classification & AUROC\\
Estrogen\cite{chembl}        & 3122    & 2-task classification & AUROC\\
MetStab\cite{metstab}         & 2267    & 2-task classification & AUROC\\
SIDER\cite{wu2018moleculenet}           & 1427    & 27-task classification & AUROC\\ 
Tox21\cite{wu2018moleculenet}           & 7831    & 12-task classification & AUROC\\
ToxCast\cite{wu2018moleculenet}         & 8575    & 617-task classification & AUROC\\ \midrule
ESOL\cite{wu2018moleculenet}            & 1128    & regression & RMSE\\
FreeSolv\cite{wu2018moleculenet}        & 642     & regression & RMSE\\
Lipophilicity\cite{wu2018moleculenet}   & 4200    & regression & RMSE\\ \midrule
ogbg-molhiv\cite{hu2020ogb}     & 41127   & binary classification & AUROC\\
ogbg-molpcba\cite{hu2020ogb}    & 437929  & 128-task classification & Average Precision\\
PCQM4Mv2\cite{hu2021ogblsc}        & 3746620 & regression & Mean Abs. Error\\ \bottomrule
\end{tabular}%
}
\end{table}

\begin{figure}[htbp]
  \centering
  \includegraphics[width=\columnwidth]{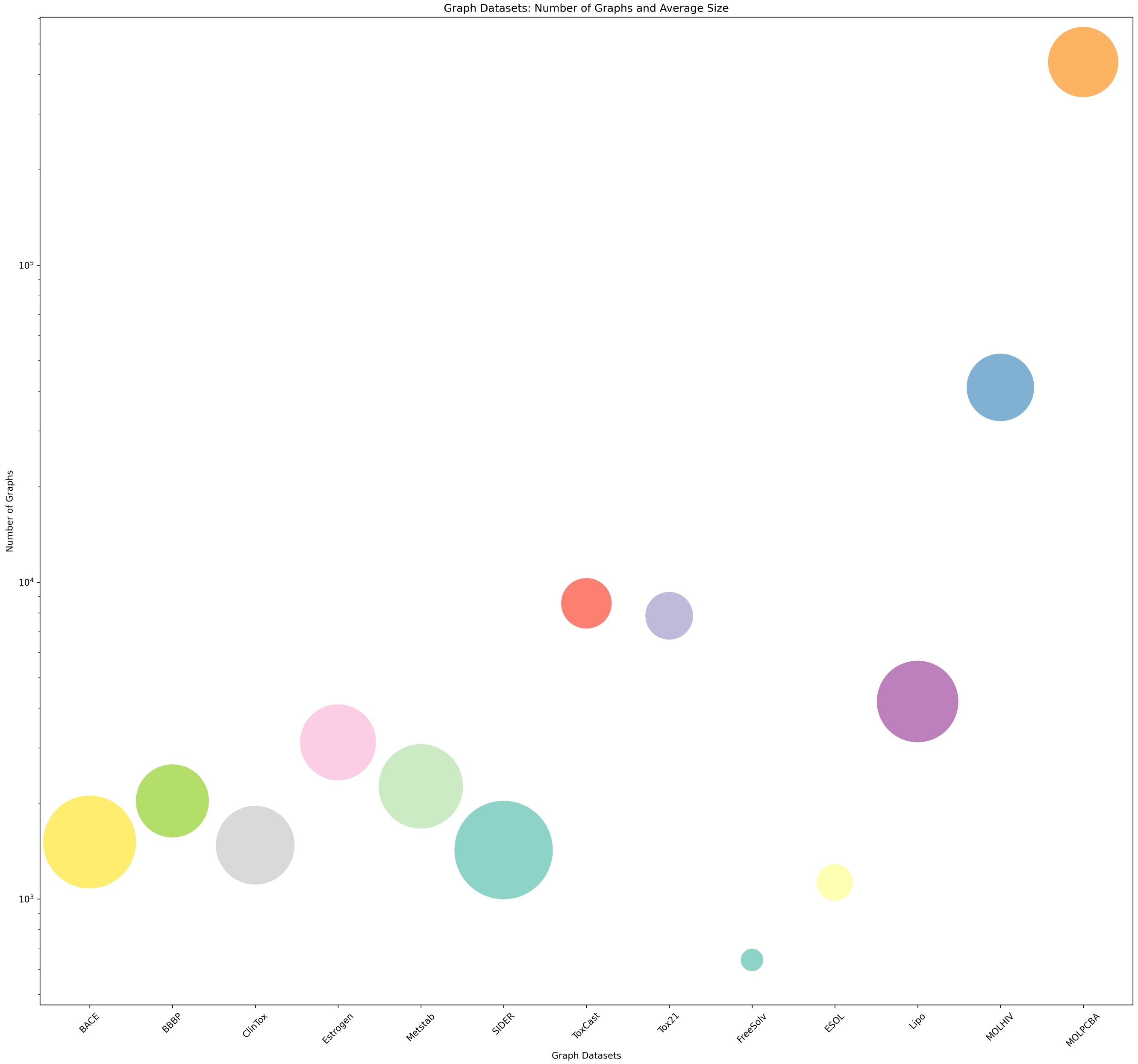}
  \caption{Overview of datasets}
  \label{fig:datasets}
\end{figure}

The summary of the molecular property datasets utilized in our experiments can be found in Table \ref{tab:datasets}. Below, you will find a breakdown of the specifics for each individual dataset:

\subsubsection{BACE}
is a dataset comprising molecules that act as inhibitors of human $\beta$-secretase 1 (BACE-1) \cite{wu2018moleculenet}.

\subsubsection{BBBP}
records whether a molecule can penetrate the blood-brain barrier \cite{wu2018moleculenet}.

\subsubsection{ClinTox}
consists of drugs that have been approved by the Food and Drug Administration (FDA) but were subsequently removed from clinical trials due to their toxic effects.

\subsubsection{Estrogen}
contains molecules with known activities towards the estrogen receptors, extracted from the ChEMBL dataset \cite{chembl}.

\subsubsection{MetStab} 
measures the half-lifetime of molecules within an organism \cite{metstab}.

\subsubsection{SIDER}
records adverse drug reactions of marketed drugs grouped into 27 system organ classes \cite{wu2018moleculenet}. 

\subsubsection{ToxCast} 
contains multiple toxicity labels for molecules obtained through high-throughput screening tests \cite{wu2018moleculenet}.

\subsubsection{Tox21}
Tox21 is a public database containing the toxicity of compounds \cite{wu2018moleculenet}.

\subsubsection{ESOL}
ESOL records the solubility of compounds \cite{wu2018moleculenet}.

\subsubsection{Lipophilicity}
Lipophilicity measures the molecular membrane permeability and solubility \cite{wu2018moleculenet}.

\subsubsection{FreeSolv}
FreeSolv contains the hydration free energy of small molecules in water from both experiments and alchemical free energy calculation \cite{wu2018moleculenet}.

\subsubsection{OGBG-MolHIV and OGBG-MolPCBA}
The datasets ogbg-molhiv and ogbg-molpcba \cite{hu2020ogb} (MIT License) are molecular property prediction datasets utilized by OGB, which were originally sourced from MoleculeNet. TThe primary objective of ogbg-molhiv is the binary classification of a molecule's suitability to inhibit HIV replication. On the other hand, ogbg-molpcba, derived from PubChem  BioAssay, is designed to predict the outcomes of 128 bioassays in a multi-task binary classification setting.

\subsubsection{PCQM4Mv2}
is considered one of the largest graph prediction datasets, comprising over 3.7 million molecules. Its primary objective is to utilize the 2-D graphs of molecules to predict their HOMO-LUMO energy gap in electronvolts (eV).

\subsection{Dataset Splits}
The main benchmarking results were obtained using 5-fold cross-validation runs, except for pretraining on PCQM4Mv2, which was conducted in a single run using the standard train/valid/test split. We ensured consistent data splitting for cross-validation and used the same seeds for models and tuning methods in all experiments. Furthermore, for ablations studies we performed replication experiments from scratch, and the results previously presented in the main text were not reused.

\subsection{Hyperparameters}
In our experiments, we conduct hyperparameter tuning for several parameters. Due to the large total number of hyperparameters, we avoid using a grid-search scheme for tuning. We adopt the AdamW optimizer with a linear "warm-up" strategy, gradually increasing the learning rate at the start of training, followed by either cosine or linear decay. The specific values for the warm-up period, base learning rate, and total number of epochs are adjusted individually for each dataset and are provided in Tables \ref{tab:hyperparameters-graphormer}, \ref{tab:hyperparameters-gps-small}, \ref{tab:hyperparameters-gps-large}, \ref{tab:hyperparameters-light} along with other relevant hyperparameters. The size of modles 

\begin{table}[htbp]
\centering
\caption{The pre-training and tuning hyperparameters for Graphormer\cite{ying2021graphormer}.}
\label{tab:hyperparameters-graphormer}
\resizebox{\columnwidth}{!}{%
    \begin{tabular}{@{}lcc@{}}
    \toprule
    Hyperparameter & Pre-training & Tuning \\
    \cmidrule{1-3}
    \# Layers & 12 & 12 \\
    \# Attention Heads & 32 & 32 \\
    Hidden Dim & 768 & 768 \\
    Dropout & 0.1 & \{0, 0.1\} \\
    Learning Rate & 2e-4 & \{ 3e-4, 1r-4\} \\
    Learning Rate Decay & polynomial & linear \\
    Weight Decay & 1e-6 & \{0, 1e-6, 1e-5, 1e-4, 1e-3, 1e-2, 1e-1, 1e1, 2e1\} \\
    Gradient Clipping & 5.0 & 5.0 \\
    AdamW $\beta$ & (0.9, 0.999) & (0.9, 0.999) \\
    Prefix Length  & - & $\{10, 20, 30 \dots 110\}$ \\
    Epochs & \{100, 200\} & 100 \\
    \bottomrule
\end{tabular}%
}
\end{table}

\begin{table}[htbp]
    \centering
    \caption{The pre-training and tuning hyperparameters for LiGhT\cite{han2022kpgt}.}
    \label{tab:hyperparameters-light}
    \resizebox{\columnwidth}{!}{%
    \begin{tabular}{@{}lcc@{}}
        \toprule
        Hyperparameter & Pre-training & Tuning \\
        \cmidrule{1-3}
        \# Layers & 12 & 12 \\
        \# Attention Heads & 12 & 12 \\
        Hidden Dim & 768 & 768 \\
        Dropout & 0.1 & \{0, 0.1\} \\
        Learning Rate & 2e-4 & \{ 3e-4, 1r-4\} \\
        Learning Rate Decay & polynomial & linear \\
        Weight Decay & 1e-6 & \{0, 1e-6, 1e-5, 1e-4, 1e-3, 1e-2, 1e-1, 1e1, 2e1\} \\
        Gradient Clipping & 5.0 & 5.0 \\
        AdamW $\beta$ & (0.9, 0.999) & (0.9, 0.999) \\
        Prefix Length  & - & $\{10, 20, 30 \dots 110\}$ \\
        Epochs & \{100, 200\} & 100 \\
        \bottomrule
    \end{tabular}%
    }
\end{table}

\begin{table}[htbp]
    \centering
    \caption{The pre-training and tuning hyperparameters for GraphGPS small\cite{ladislav2022graphgps}.}
    \label{tab:hyperparameters-gps-small}
    \resizebox{\columnwidth}{!}{%
        \begin{tabular}{@{}lcc@{}}
        \toprule
        Hyperparameter & Pre-training & Tuning \\
        \cmidrule{1-3}
        \# Layers & 16 & 16 \\
        \# Attention Heads & 8 & 8 \\
        Hidden Dim & 256 & 256 \\
        Dropout & 0.1 & \{0, 0.1\} \\
        Learning Rate & 2e-4 & \{ 1e-3, 3e-4, 1r-4\} \\
        Learning Rate Decay & polynomial & cosine with warm-up \\
        Learning Rate Warm-up Epochs & - & 5 \\
        Weight Decay & 1e-6 & \{0, 1e-6, 1e-5, 1e-4\} \\
        Gradient Clipping & 5.0 & 5.0 \\
        AdamW $\beta$ & (0.9, 0.999) & (0.9, 0.999) \\
        Prefix Length  & - & $\{10, 20, 30 \dots 110\}$ \\
        Epochs & \{100, 200\} & 100 \\
        \bottomrule
    \end{tabular}%
    }
\end{table}

\begin{table}[htbp]
\centering
\caption{The pre-training and tuning hyperparameters for GraphGPS large\cite{ladislav2022graphgps}.}
\label{tab:hyperparameters-gps-large}
\resizebox{\columnwidth}{!}{%
\begin{tabular}{@{}lcc@{}}
    \toprule
    Hyperparameter & Pre-training & Tuning \\
    \cmidrule{1-3}
    \# Layers & 12 & 12 \\
    \# Attention Heads & 16 & 16 \\
    Hidden Dim & 768 & 768 \\
    Dropout & 0.1 & \{0, 0.1\} \\
    Learning Rate & 2e-4 & \{ 1e-3, 3e-4, 1r-4\} \\
    Learning Rate Decay & polynomial & cosine with warm-up \\
    Learning Rate Warm-up Epochs & - & 5 \\
    Weight Decay & 1e-6 & \{0, 1e-6, 1e-5, 1e-4\} \\
    Gradient Clipping & 5.0 & 5.0 \\
    AdamW $\beta$ & (0.9, 0.999) & (0.9, 0.999) \\
    Prefix Length  & - & $\{10, 20, 30 \dots 110\}$ \\
    Epochs & \{100, 200\} & 100 \\
    \bottomrule
\end{tabular}%
}
\end{table}

\begin{table}[htbp]
  \centering
  \caption{Graph Transformer architectures and number of parameters.}
  \label{tab:gt-sizes}
  \begin{tabular}{c|c}
    \hline
    Model    & Size \\
    \hline
    GraphGPS Small \cite{ladislav2022graphgps} & 14M\\
    \hline
    Graphormer \cite{ying2021graphormer} & 48M\\
    \hline
    LiGhT \cite{han2022kpgt} & 90M\\
    \hline
    GraphGPS Large \cite{ladislav2022graphgps} & 103M\\
    \hline
  \end{tabular}
\end{table}

\section{More Ablation Studies}
\begin{table}[htbp]
\centering
\caption{Ablation studies of each component. }
\label{tab:moleculenet-regression-ablations}
\resizebox{\columnwidth}{!}{%
\begin{tabular}{@{}lccc@{}}
\toprule
\multirow{2}{*}{Model}  & \multicolumn{3}{c}{Moleculenet Regression Dataset ($\downarrow$)} \\ \cmidrule(l){2-4} 
                     & ESOL & FreeSolv & Lipo \\ \midrule
Graphormer  Lightweight Tuning        & 1.123 ± 0.044 & 3.085 ± 0.378 & 0.915 ± 0.032  \\
Graphormer  Deep Prefix Tuning        & 0.910 ± 0.041 & 1.657 ± 0.351 & 0.863 ± 0.026  \\
Graphormer DeepGPT                    & 0.943 ± 0.024 & 1.668 ± 0.114  & 0.834 ± 0.046  \\ \midrule
GraphGPS Lightweight Tuning           & 2.306 ± 0.419 & 4.793 ± 3.652 & 1.043 ± 0.091  \\
GraphGPS Deep Prefix Tuning           & 0.589 ± 0.087 & 2.221 ± 0.177 & 0.459 ± 0.040  \\
GraphGPS DeepGPT                      & 0.685 ± 0.130 & 1.415 ± 0.254 & 0.528 ± 0.056  \\ \bottomrule

\end{tabular}%
}
\end{table}

\begin{table*}[htbp]
\centering
\caption{Ablation studies of each component}
\label{tab:ablation-tune-comple}
\resizebox{\textwidth}{!}{%
\begin{tabular}{@{}lccccccccc@{}}
\toprule
\multirow{2}{*}{Model} & \multirow{2}{*}{\#Params} & \multicolumn{8}{c}{Classification Dataset ($\uparrow$)}               \\ \cmidrule(l){3-10} 
                            &                      & BACE & BBBP$^\ast$ & ClinTox$^\ast$ & Estrogen$^\ast$ & MetStab$^\ast$ & SIDER & Tox21 & ToxCast$^\ast$ \\ \midrule
Graphormer Lightweight Tuning &  & 0.847 ± 0.013 & 0.882 ± 0.016 & 0.766 ± 0.082 & 0.935 ± 0.015 & 0.863 ± 0.021 & 0.625 ± 0.017 & 0.797 ± 0.008 & 0.705 ± 0.011\\ 
Graphormer Deep Prefix Tuning &  & 0.863 ± 0.013 & 0.904 ± 0.013 & 0.860 ± 0.062 & 0.937 ± 0.016 & 0.883 ± 0.014 & 0.584 ± 0.009 & 0.790 ± 0.010 & 0.705 ± 0.015\\ 
Graphormer DeepGPT &  & 0.883 ± 0.019 & 0.914 ± 0.016 & 0.884 ± 0.030 & 0.941 ± 0.006 & 0.871 ± 0.004 & 0.646 ± 0.003 & 0.813 ± 0.010 & 0.725 ± 0.011\\ \midrule
GraphGPS Lightweight Tuning  &   & 0.819 ± 0.010 & 0.751 ± 0.041 & 0.692 ± 0.107 & 0.877 ± 0.012 & 0.673 ± 0.031 & 0.541 ± 0.015 & 0.667 ± 0.008 & 0.519 ± 0.005\\
GraphGPS Deep Prefix Tuning  &   & 0.854 ± 0.016 & 0.835 ± 0.030 & 0.866 ± 0.059 & 0.902 ± 0.014 & 0.841 ± 0.014 & 0.598 ± 0.019 & 0.821 ± 0.014 & 0.690 ± 0.009\\
GraphGPS DeepGPT &    & 0.892 ± 0.022 & 0.901 ± 0.015 & 0.907 ± 0.044 & 0.944 ± 0.010 & 0.899 ± 0.019 & 0.609 ± 0.020 & 0.832 ± 0.012 & 0.735 ± 0.005\\ \bottomrule
\end{tabular}%
}
\end{table*}

\begin{figure}[htbp]
    \centering
    \includegraphics[width=0.8\columnwidth]{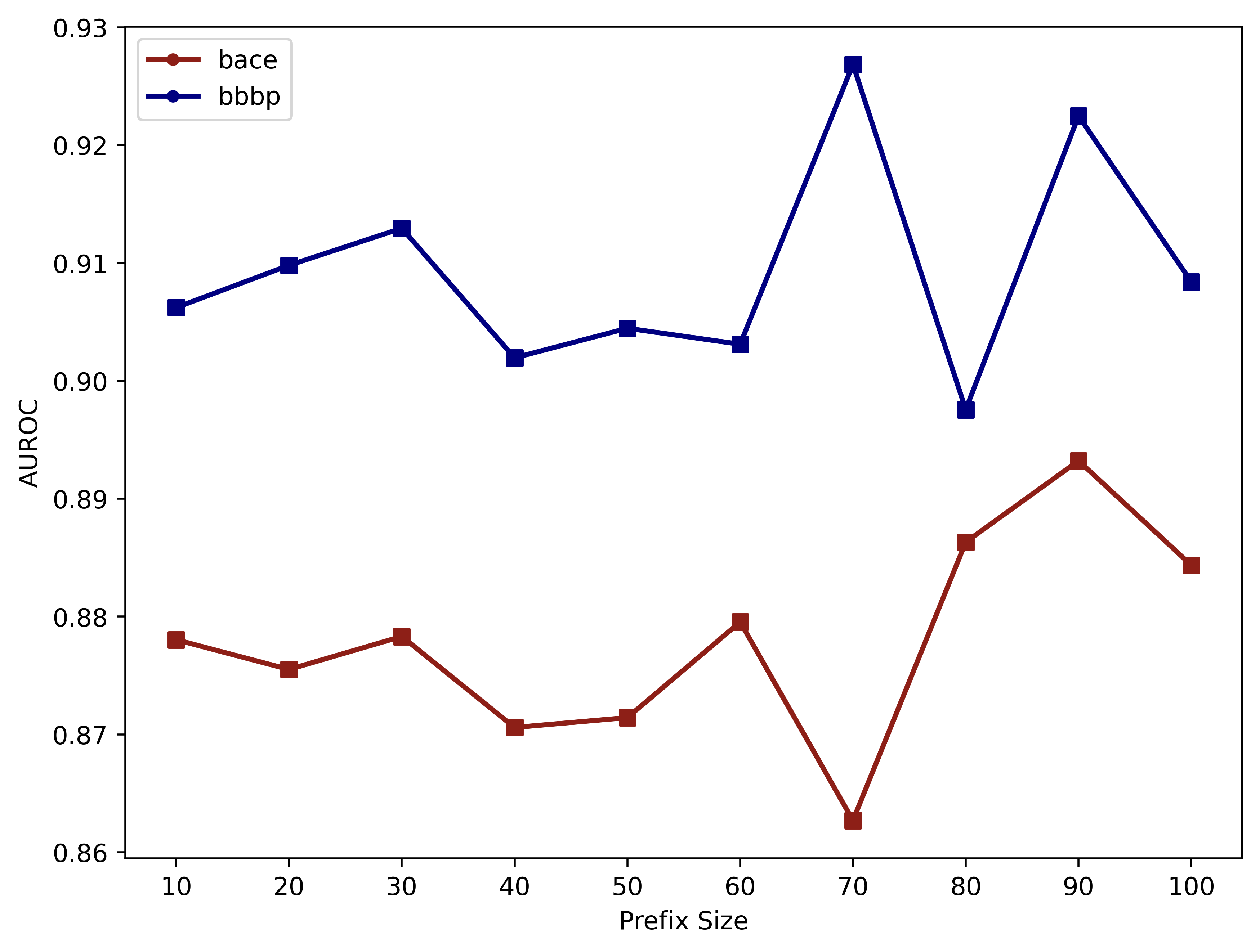}
    \caption{Effect of prompt length using Graphormer.}  
    \label{fig:ablations-size}  
\end{figure}

\begin{figure*}[htbp]
    \begin{subfigure}{0.33\textwidth}
        \centering
        \includegraphics[width=\textwidth]{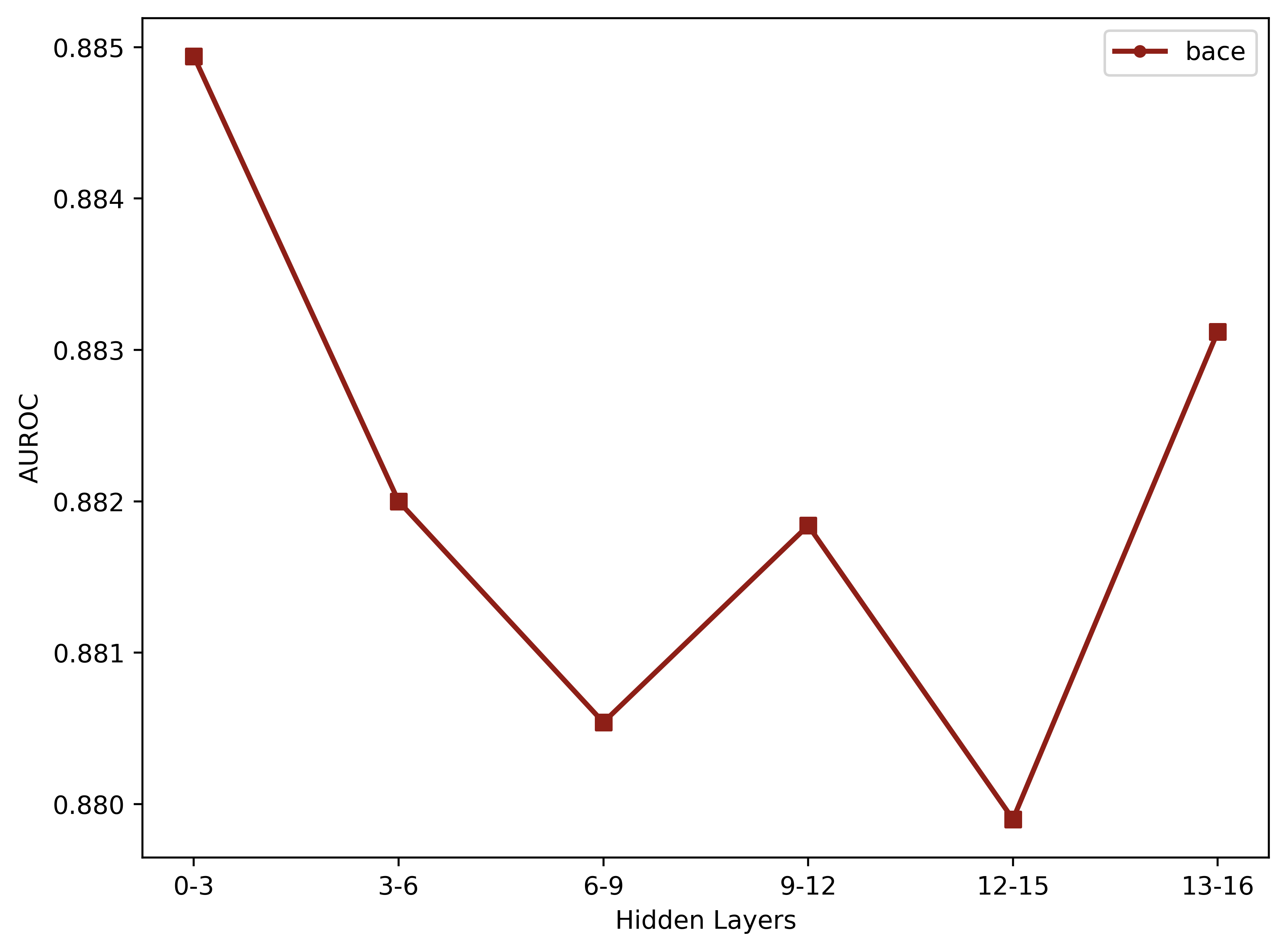}
        \caption{Fixed 3 layers}  
        \label{fig:gps-bace-ablations}  
    \end{subfigure}
    \hfill
    \begin{subfigure}{0.33\textwidth}
        \centering
        \includegraphics[width=\textwidth]{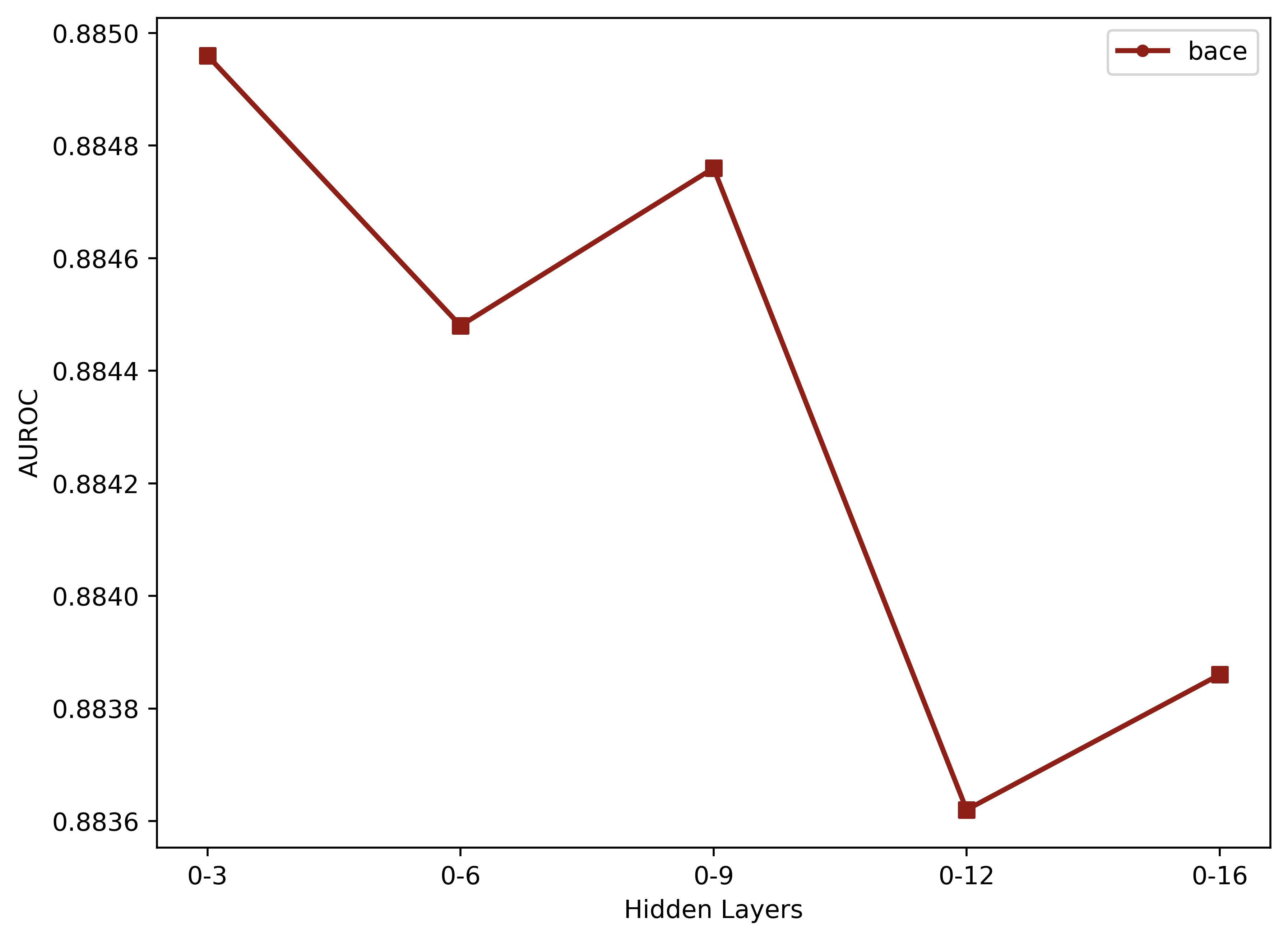}
        \caption{First k layers. }  
        \label{fig:gps-bace-ablations-asc}  
    \end{subfigure}
    \hfill
    \begin{subfigure}{0.33\textwidth}
        \centering
        \includegraphics[width=\textwidth]{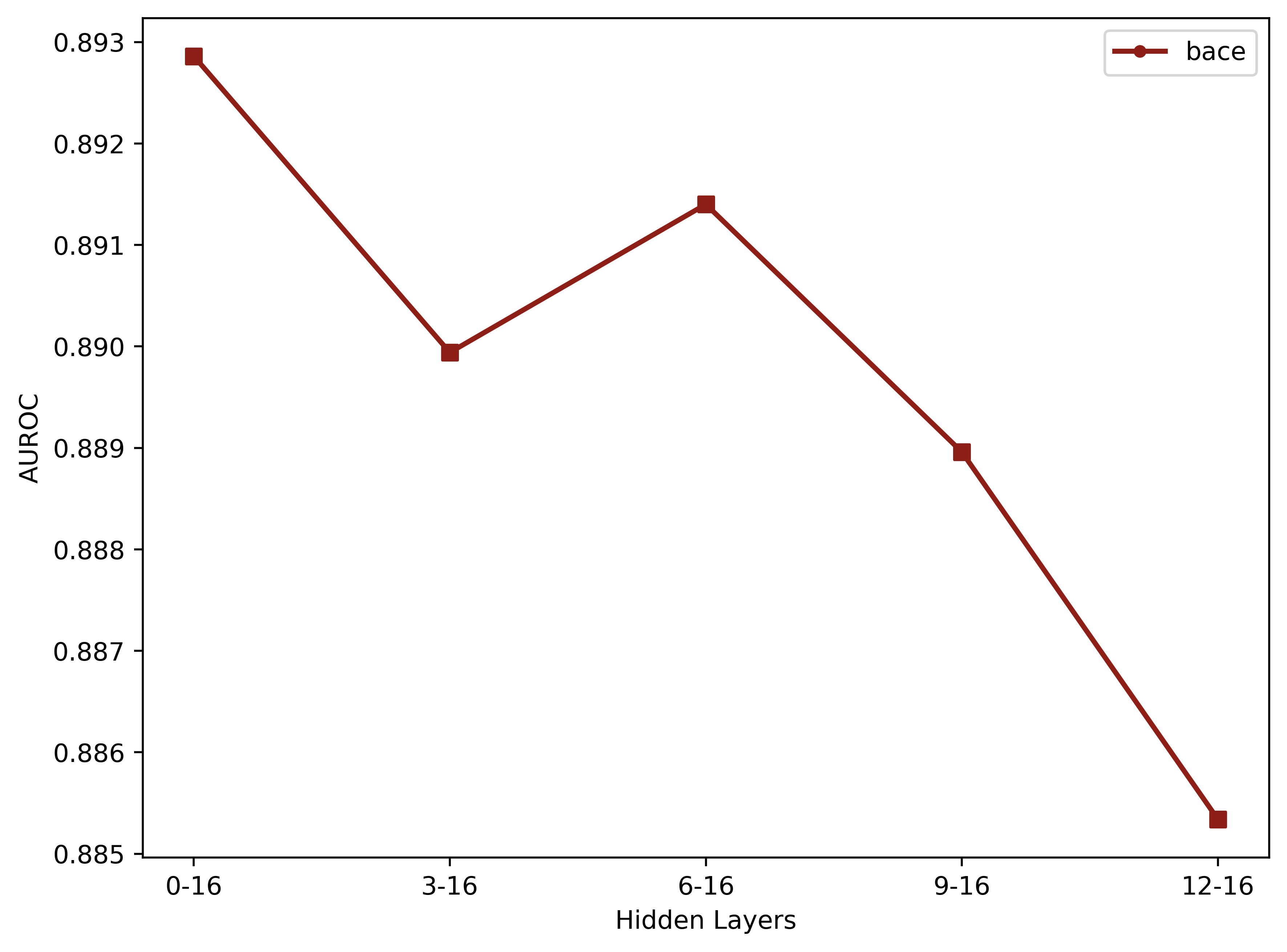}
        \caption{Last k layers.}  
        \label{fig:gps-bace-ablations-desc}  
    \end{subfigure}
    \caption{Impact of the depth of injected prompt tokens using GraphGPS on BACE dataset. The notation "[x-y]" represents the layer-interval where continuous prompts are added (e.g., "3-6" indicates prompts added to transformer layers from 3 to 6) .}
\label{fig:gps-bace}
\end{figure*}

\begin{figure*}[htbp]
  \begin{subfigure}{0.33\textwidth}
        \centering
        \includegraphics[width=\textwidth]{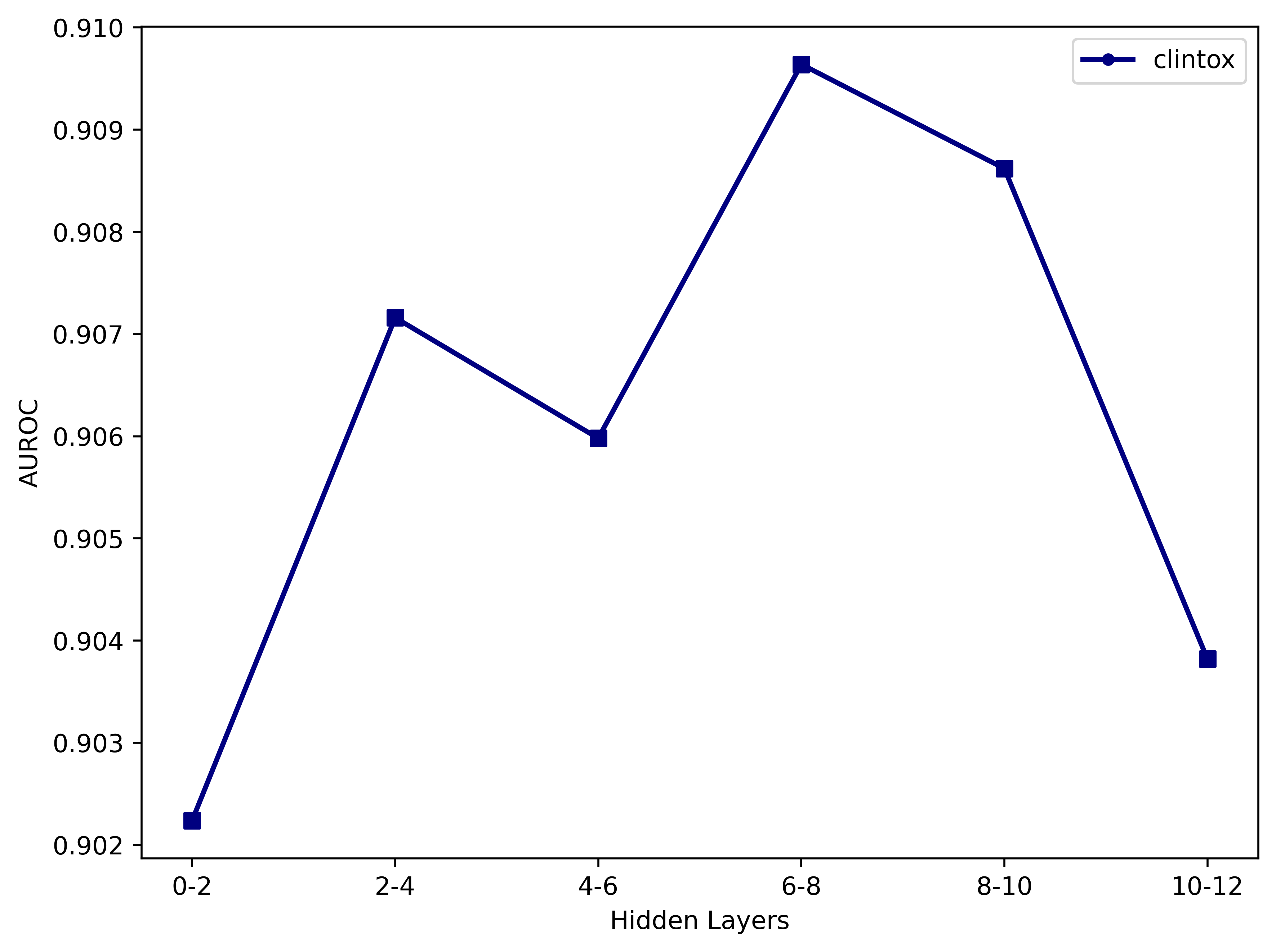}
        \caption{Fixed 2 layers}
        \label{fig:gps-clintox-depth-ablation}    
  \end{subfigure}
  \hfill
  \begin{subfigure}{0.33\textwidth}
      \centering
      \includegraphics[width=\textwidth]{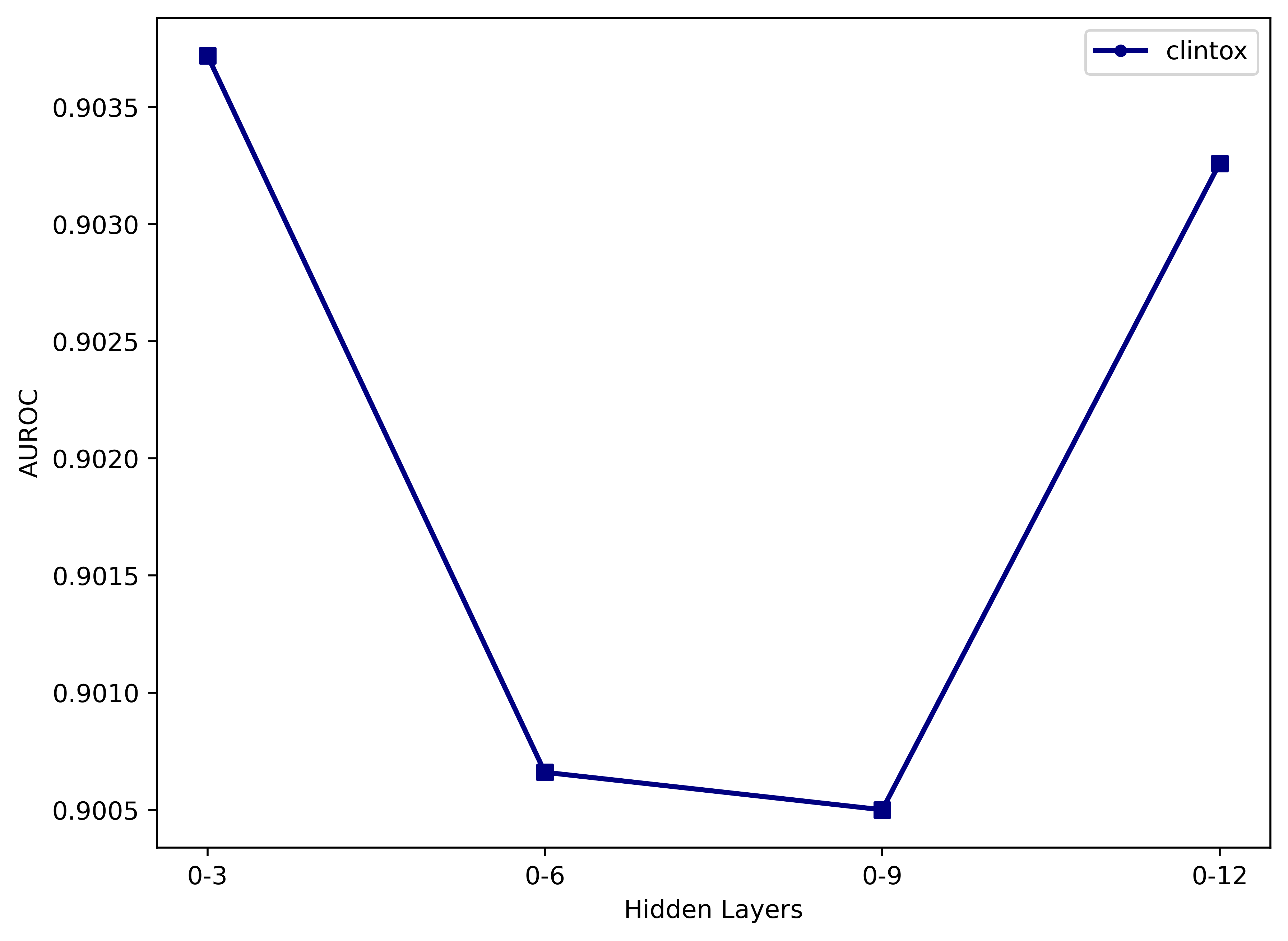}
      \caption{First k layers}
      \label{fig:gps-clintox-depth-ablation-asc}  
  \end{subfigure}
  \begin{subfigure}{0.33\textwidth}
        \centering
        \includegraphics[width=\textwidth]{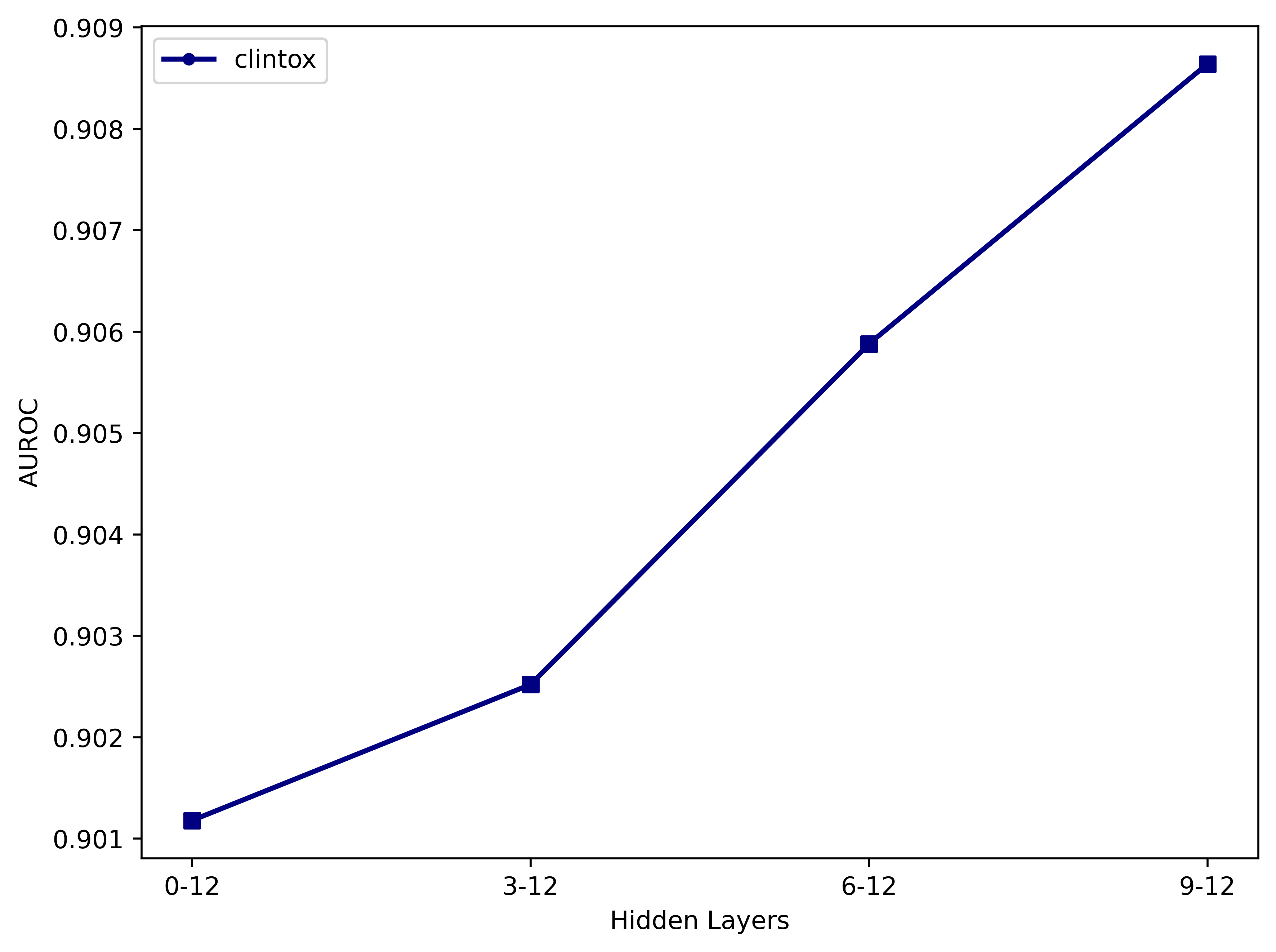}
        \caption{Last k layers}
        \label{fig:gps-clintox-depth-ablation-desc}    
  \end{subfigure}
  \caption{Impact of injecting prompt tokens to using GraphGPS on Clintox.}
\label{fig:gps-clintox-ablations}  
\end{figure*}

\end{document}